\documentclass[paper]{ieice}
\usepackage[fleqn]{amsmath}
\usepackage{newtxtext}
\usepackage[varg]{newtxmath}

\field{D}
\title{Enhanced Data Transfer Cooperating with Artificial Triplets for Scene Graph Generation}

\authorlist{%
 \authorentry[kcchu@nlab.ci.i.u-tokyo.ac.jp]{KuanChao Chu}{n}{UTokyo}\MembershipNumber{}
 \authorentry{Satoshi Yamazaki}{n}{NEC}\MembershipNumber{}
 \authorentry{Hideki Nakayama}{m}{UTokyo}\MembershipNumber{0731269}
}

\affiliate[UTokyo]{The authors are with the University of Tokyo, Tokyo, 113-8654 Japan.
}
\affiliate[NEC]{The author is with the NEC Corporation, Tokyo, 108-8001 Japan.
}

\received{2023}{11}{1}
\revised{2024}{2}{27}

\usepackage{graphicx}
\usepackage{amsmath}
\usepackage{booktabs}

\usepackage{microtype}      
\usepackage{nicefrac}       
\usepackage{float}
\usepackage{url}            
\usepackage{multirow}

\usepackage[ruled, noline, linesnumbered]{algorithm2e}
\usepackage{subcaption}
\usepackage{comment}
\usepackage{xcolor}         
\definecolor{mybluegreen}{RGB}{55, 111, 113}
\definecolor{mybrown}{RGB}{204, 88, 75}

\begin{document}
\maketitle
\begin{summary}
This work focuses on training dataset enhancement of informative relational triplets for Scene Graph Generation (SGG).
Due to the lack of effective supervision, the current SGG model predictions perform poorly for informative relational triplets with inadequate training samples.
Therefore, we propose two novel training dataset enhancement modules: Feature Space Triplet Augmentation (FSTA) and Soft Transfer. FSTA leverages a feature generator trained to generate representations of an object in relational triplets.
The biased prediction based sampling in FSTA efficiently augments artificial triplets focusing on the challenging ones. In addition, we introduce Soft Transfer, which assigns soft predicate labels to general relational triplets to make more supervisions for informative predicate classes effectively.
Experimental results show that integrating FSTA and Soft Transfer achieve high levels of both Recall and mean Recall in Visual Genome dataset. The mean of Recall and mean Recall is the highest among all the existing model-agnostic methods.
\end{summary}

\section{Introduction}
\label{sec:intro}

\begin{figure}[ht]
  \centering
  \begin{subfigure}{0.48\columnwidth}
    \includegraphics[width=\linewidth]{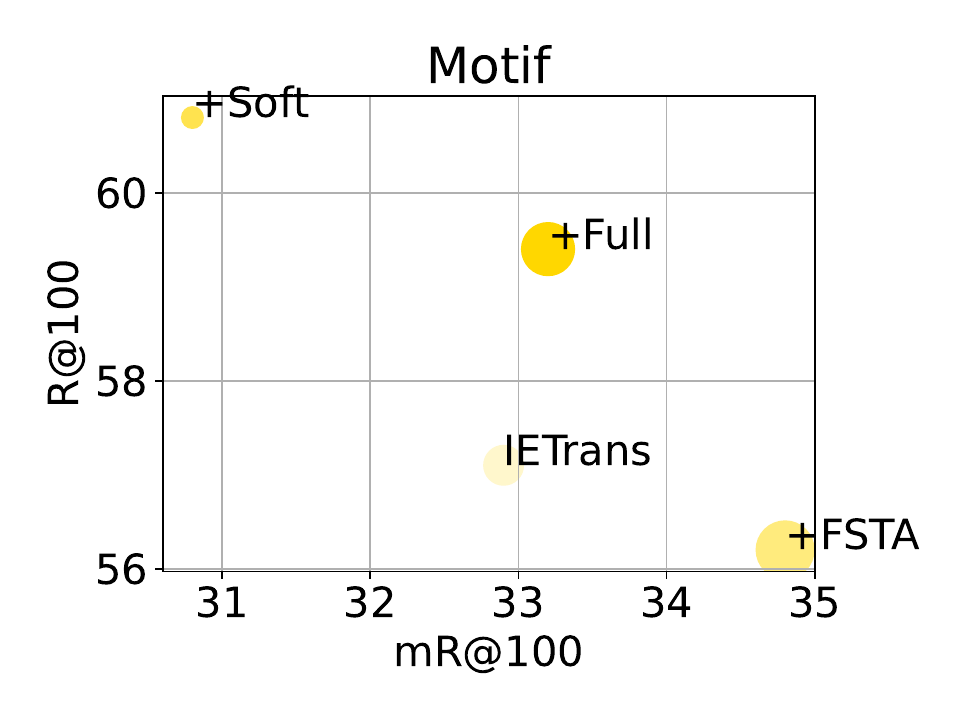}
  \end{subfigure}
  \begin{subfigure}{0.48\columnwidth}
    \includegraphics[width=\linewidth]{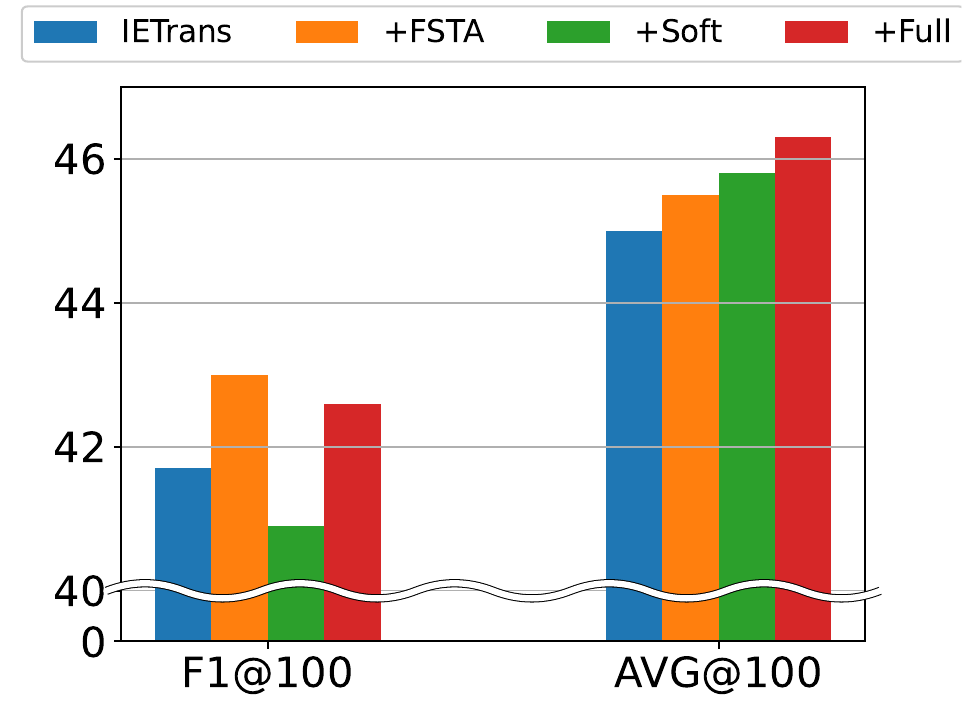}
  \end{subfigure}
  \begin{subfigure}{0.48\columnwidth}
    \includegraphics[width=\linewidth]{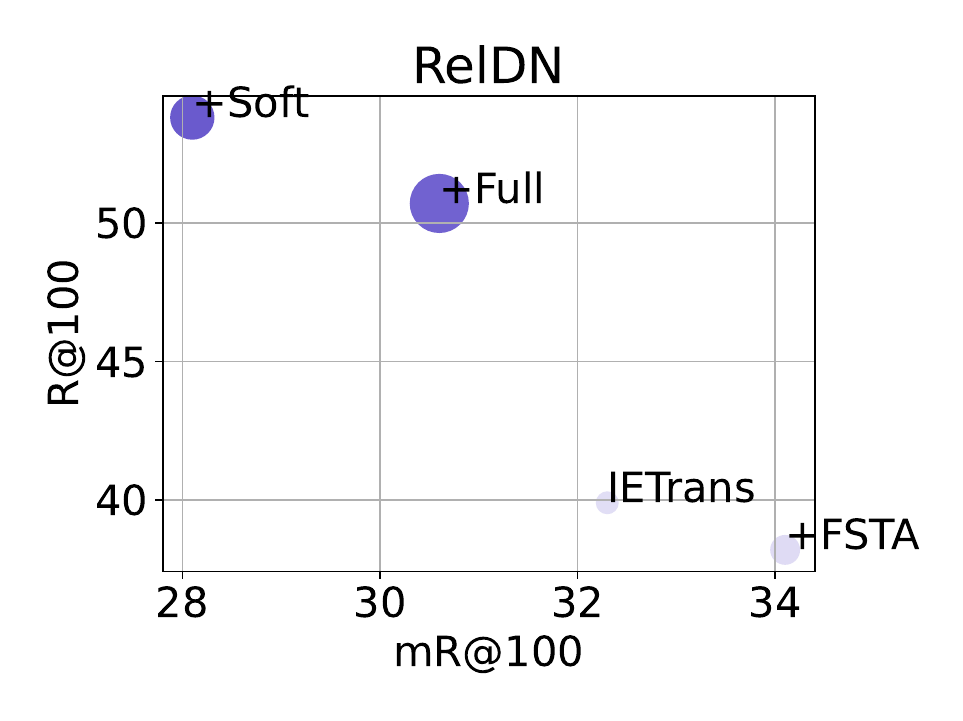}
  \end{subfigure}
  \begin{subfigure}{0.48\columnwidth}
    \includegraphics[width=\linewidth]{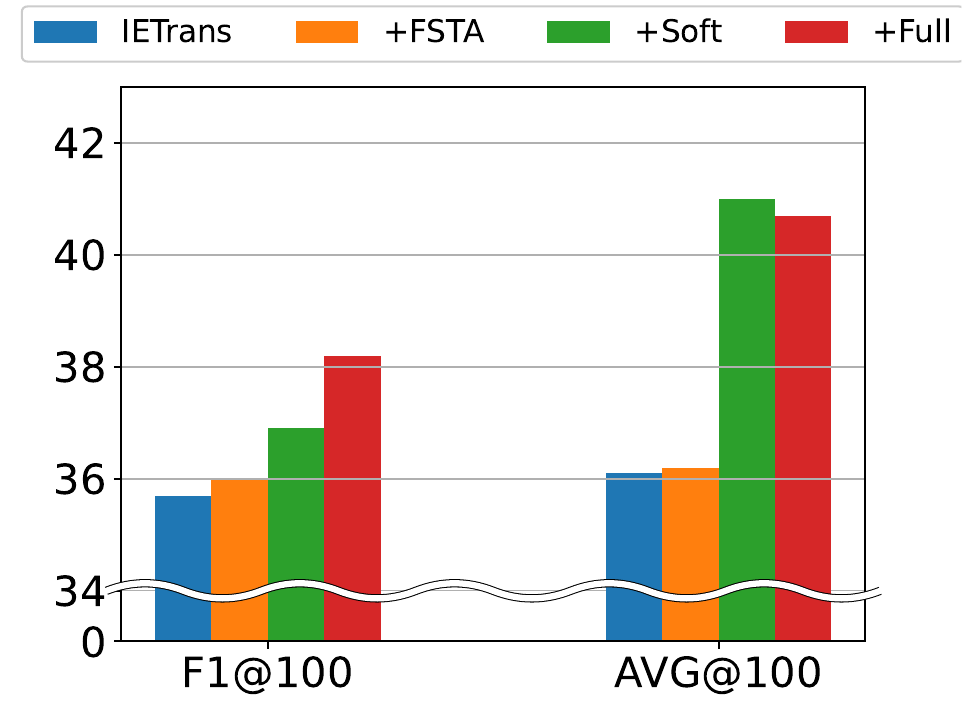}
  \end{subfigure}
  
  \caption{Accuracy comparison between FSTA, Soft Transfer, Full, and the baseline IETrans on Motif ($1^{st}$ row) and RelDN ($2^{nd}$ row). In the scatter plots (left), a larger dot size and a darker color represent higher F1@100 and AVG@100 scores, respectively. As shown in the bar plots (right), increased scores in the overall metrics (F1@100 and AVG@100) indicate the alleviated performance trade-off in our full method, consisting of two complementary modules.}
\label{fig:first image}
\end{figure}
\vspace{-3mm}

Scene graphs have emerged as a pivotal representation for detailing semantic information within a visual scene, by specifying relationships between object pairs \cite{SGorigin_IMretrieval,SGG_Survey}. This representation enables reasoning about visual content through the encoded spatial and logical details of object instances and their relations. 
In modern applications, scene graphs have become foundational for high-level visual tasks like activity parsing \cite{MOMA}, image retrieval \cite{SGorigin_IMretrieval}, visual understanding \cite{Appl_image_understanding}, and image captioning \cite{Appl_captioning}. This paper delves into the scene graph generation (SGG) task, aiming to predict objects and their relations from visual input.

SGG models encounter two primary challenges when trained on common dataset \cite{VisualGenome}: first, the distinct long-tailed distribution of relations \cite{DC2-ABCS,longtail_decouple}, and second, the ambiguity caused by semantically similar relation classes (e.g., on/on back of/mounted on) \cite{NICE,IETrans,FGPL}. The latter exacerbates the issue, as instances within a category may be annotated under multiple confusing classes. Such complexities often bias relation predictions in general SGG models, leading to low recall rates for rare predicate classes. While some unbiased SGG methods \cite{DC2-ABCS,TDE,DLFE,PCPL} have addressed this, they often sacrifices performance on frequent classes. Hence, it is essential to consider these trade-offs to ensure the model performance on the majority of data is not compromised.

Recently, the training data modification approach has shown promising results for training an unbiased SGG model \cite{NICE,IETrans}. Two major concepts for the modification
are the addition of new predicate labels and reassignment of existing ones, which can efficiently improve rare class performance. We revisit these concepts through IETrans \cite{IETrans}, a baseline data modification method. IETrans encompasses two modules: external transfer for label addition and internal transfer for label reassignment. Notably, the external transfer, while leveraging background triplets for augmentation, doesn't fully exploit the available data. Given the compositional nature of relational triplets, inter-triplet augmentation appears worthwhile. Additionally, predicate reassignments in the internal transfer are not uniformly reliable. A human evaluation study \cite{IETrans} reveals that only 76\% of transferred triplets are deemed reliable. The inconsistency in the degree of semantic confusion, even among identical predicates, suggests that an ``entire'' transfer strategy might not be optimal. Guided by these findings, our system seeks to address these shortcomings by extending the modification concepts in two key ways: improving upon the data addition process and enhancing the reassignment efficiency. 

Our method introduces two novel modules: Feature Space Triplet Augmentation (FSTA) and Soft Transfer. FSTA dynamically creates artificial triplets during training. We can construct new data by enumerating triplet combinations \verb+subject-predicate-object'+ and \verb+subject'-predicate-object+ from a sampled mini-batch. Here, $x'$ denotes data not from the original triplets. These artificial triplets serve to regularize the relation classification module in the SGG model. We undersample the frequent classes in artificial triplets to shape their predicate distribution. Further, a biased prediction-based sampler selects the class label for $x'$. This design aims to often sample combinations that are hard to be predicted correctly for a biased model. A pre-trained generator synthesizes the corresponding features based on class labels. 
Besides, Soft Transfer refines label reassignment by implementing an instance-wise ranking and mapping mechanism. We first compute a reliability score for each reassigned sample from biased model predictions, then select low-scoring triplets for Soft Transfer. Subsequently, a non-binary predicate label is calculated by mapping the reliability score, allowing for finer control over semantic confusion by using this label probability instead of an entire reassignment.  

FSTA notably boosts performance on rare classes with increased sample quantity and diversity. Conversely, Soft Transfer alleviates performance loss in frequent classes, a typical compromise when elevating rare class performances. In essence, while FSTA contributes to mean recall (mR) gain, Soft Transfer leads to the recall (R) gain. Collectively, these modules bring reduced performance trade-off that shown in the improved overall metrics, F1@K and Avg@K. Our model-agnostic method was evaluated on the VisualGenome dataset \cite{VisualGenome}, using two types of general SGG models: MOTIF \cite{Motif} and RelDN \cite{RelDN} with IETrans. In the predcls task, our system outperforms the baseline IETrans by a 3.1\% and 7.0\% relative gain on the F1@100 metric for MOTIF and RelDN, respectively. Fig. \ref{fig:first image} illustrates the balanced performance of our method.

To sum up, we make the following \textit{contributions}:
\begin{enumerate}
    \item We propose a novel, model-agnostic method for training a R/mR balanced SGG model. It integrates two complementary modules: FSTA and Soft Transfer, which enhance the baseline IETrans.  
    
    \item We conduct extensive experiments and discussions on VisualGenome and demonstrate the effectiveness of our system. 
\end{enumerate}

\begin{figure*}[t]
  \centering
  \includegraphics[trim={2.31cm 4.05cm 1.50cm 3.03cm},clip,width=\linewidth]{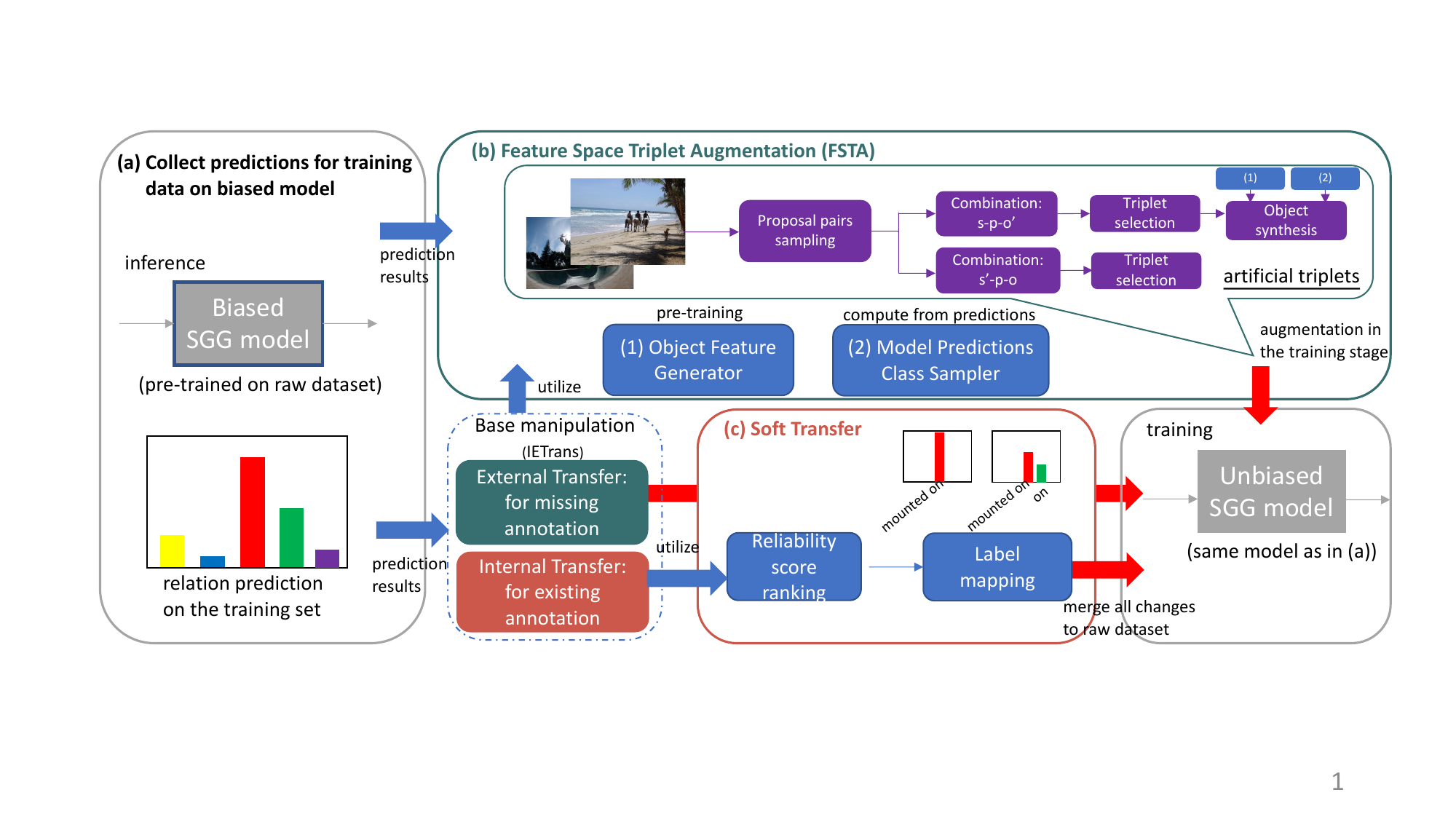}
  \caption{The system overview of our proposed method. The 
  \textcolor{mybluegreen}{FSTA} and \textcolor{mybrown}{Soft Transfer} modules are designed to introduce new concepts to enhance the baseline dataset manipulation module, IETrans. Blocks indicated in blue are prepared during the pre-processing stage, whereas the blocks in purple are designated for the unbiased SGG model training stage.}
  \label{sys_overview}
\end{figure*}
\vspace{-3mm}

\section{Related Work}
\subsection{Biased and Unbiased Scene Graph Generation}
Scene Graph Generation (SGG) is first proposed as visual relation detection (VRD) \cite{lu2016visual}, where each relation is independently detected, ignoring the rich contextual information.
Later studies in SGG utilizes advanced techniques, e.g., message passing \cite{xu2017scene}, recurrent sequential architectures \cite{Motif} or contrastive learning \cite{RelDN}. 
However the accuracy of relationship detection is far from satisfaction due to the heavily biased data.
Some authors \cite{VCTree,chen2019knowledge} point out that the predictions of current SGG models often collapse to several general and trivial predicate classes.
Instead of only focusing on recall metric, 
hence they propose a new metric named mean recall, which is the average recall of all predicate classes, as the unbiased metric.
Efforts towards developing unbiased SGG models have been noted.
BGNN and DT2-ACBS \cite{BGNN,DC2-ABCS} proposes sophisticated re-sampling strategy.
Some debiasing solutions \cite{TDE,yu2021cogtree,IETrans} are categorized as biased-model based strategies that utilize predictions from biased SGG model.
Especially, IETrans\cite{IETrans} adopts triplet-level data transfers over the less precise predicate-level manipulation.
Our proposed method is inspired from IETrans, and focuses on data augmentation for inadequate training samples.

\subsection{Compositional Learning}
Recognition-By-Components theory \cite{biederman1987recognition} which illustrates that human representations of concepts are decomposable is especially influential in object recognition.
Based on the theory, novel concepts from a few samples can be potentially learnable by composing known primitives.
Some authors apply the compositional deep representation into few-shot learning for object recognition \cite{tokmakov2019learning} and Human-Object Interaction (HOI) detection \cite{kato2018compositional,VCL,FCL}. 
Visual compositional learning frameworks \cite{VCL,FCL} proposed for HOI detection compose HOI training samples from image-pairs and fake object representations to solve the open long-tail issue in HOI detection.
Our proposed data augmentation method adapts the compositional learning to SGG task.
To overcome the biased data issue in SGG, our sampling strategy of composed training samples plays an important role.

\section{Methodology}

A scene graph generation (SGG) model predicts a direct graph $G$ for an input image $I \in \mathbb{R}^3$. 
$G = \lbrace V, E\rbrace$ contains a set of predicted objects $V = \lbrace(\mathbf{b}_{i}, c_{e_{i}})\}^{N_{V}}_{i=1}$ and a set of predicted relationships $E = \lbrace(s_{j}, c_{r_{j}}, o_{j})\}^{N_{E}}_{j=1}$. 
$\mathbf{b}_{i} \in \mathbb{R}^4$ denotes the position of an object using bounded box coordinates. 
$c_{e_{i}} \in \mathcal{C}_{objects}$ and $c_{r_{j}} \in \mathcal{C}_{relations}$ belong to the known object and relation classes, respectively.
$s_{j} \in V$ and $o_{j} \in V$ are nodes connected by the relation $c_{r_{j}}$. Each element in $E$ can also be depicted as a \emph{subject-predicate-object} triplet to convey the intrinsic semantic information.

Conceptually, an SGG model can be seen as a sequence of modules that includes an object detection backbone followed by a relation prediction head. The detection backbone first outputs a set of Region of Interest (RoIs) containing the detected object information. These results are then forwarded to the relation prediction head to refine these detections and predict the relation between the RoI pairs. This work mainly focuses on the scenario where a maximum of one predicate, with the highest score, can be predicted between each RoI pair. This principle aligns with the \textit{graph constraint} mode described in other research. 
 
Figure \ref{sys_overview} illustrates the system overview. In our enhanced data modification approach, we further leverage predictions from a pre-trained yet biased SGG model. Sec. \ref{ietrans-intro} gives a brief overview of the baseline modification, IETrans. Sec. \ref{fsta} details our FSTA module, elucidating a strategy for triplet augmentation in the feature space during the unbiased \textbf{training phase}. Sec. \ref{softlabel} explains Soft Transfer, a method offering precise control over reassigning predicate labels during the \textbf{pre-processing stage}, ensuring better handling of per-sample semantic confusion. Sec. \ref{3_5_implementation} has our implementation details.

\subsection{Preliminary Introduction: IETrans}
\label{ietrans-intro}

IETrans constructs a modified training dataset during pre-processing. It has two steps: external transfer and internal transfer. 
In the external transfer, it acquires new labels from those no-relation object pairs. By ranking these no-relation prediction scores from the biased model, some object pairs are assigned new predicate labels that have the highest probability. This approach, however, may not fully leverage the available data.  
On the other hand, internal transfer shifts general predicates to informative ones using a ranking and affinity score filtering method through biased prediction results. For example, ``man-on-horse'' becomes ``man-sitting on-horse''. Nevertheless, the level of ambiguity is context-sensitive, and a binary transfer decision might not effectively capture semantic confusion across all samples. 

These transfer steps explicitly adjust the balance of the dataset distribution, based on the property that a rare predicate class is often a more informative version of a frequent class. 
Linking general to informative predicate pairs reduced semantic ambiguity by discovering the confusion in biased model predictions. 
The raw frequently prior is employed to compensate the largely sacrificed performance on general predicates. We refer readers to the original publication for more details.

\subsection{Feature Space Triplet Augmentation}
\label{fsta}

Given the compositional nature of a relation triplet, it's possible to construct a new sample from multiple existing ones. The interaction between object-predicate representations in the feature space for SGG models is pivotal. Even though they can be combined in various ways—be it addition \cite{RelDN}, concatenation \cite{RelDN}, or element-wise multiplication \cite{Motif}—the upstream feature extractor processes the elements in a triplet independently. As such, when the object representation in a triplet is partially changed to form a new semantically reasonable combination, the relation predictor in the relation head should be encouraged to produce similar outputs. This can be represented as:
\begin{equation}
\begin{aligned}
  M(F(\mathbf{f}_{s_{i}}, \mathbf{f}_{p_{i}}, \mathbf{f}_{o_{i}}; &\theta_{F});\theta_{M}) \approx\\ 
  &M(F(\mathbf{f}_{s_{i}}, \mathbf{f}_{p_{i}}, \mathbf{f}_{o_{j}}; \theta_{F});\theta_{M})
\end{aligned}  
\end{equation}
where $(\mathbf{f}_{s_{i}}, \mathbf{f}_{p_{i}}, \mathbf{f}_{o_{i}})$ denotes the subject-relation-object intermediate representations of the $i^{th}$ sample. Given that $i\neq j$ and $(c_{s_{i}}, c_{p_{i}}, c_{o_{j}})$ is a semantically reasonable triplet, $M(\cdot;\theta_{M})$ is the final predicate classification module, and $F(\cdot;\theta_{F})$ symbolizes the transitional layers in between. In light of this, artificial triplets can serve as augmented data to regularize the relation predictor during training. 

\begin{figure}[t]
  \centering
  \includegraphics[trim={1.6cm 1.45cm 6.45cm 2.85cm},clip,width=\linewidth]{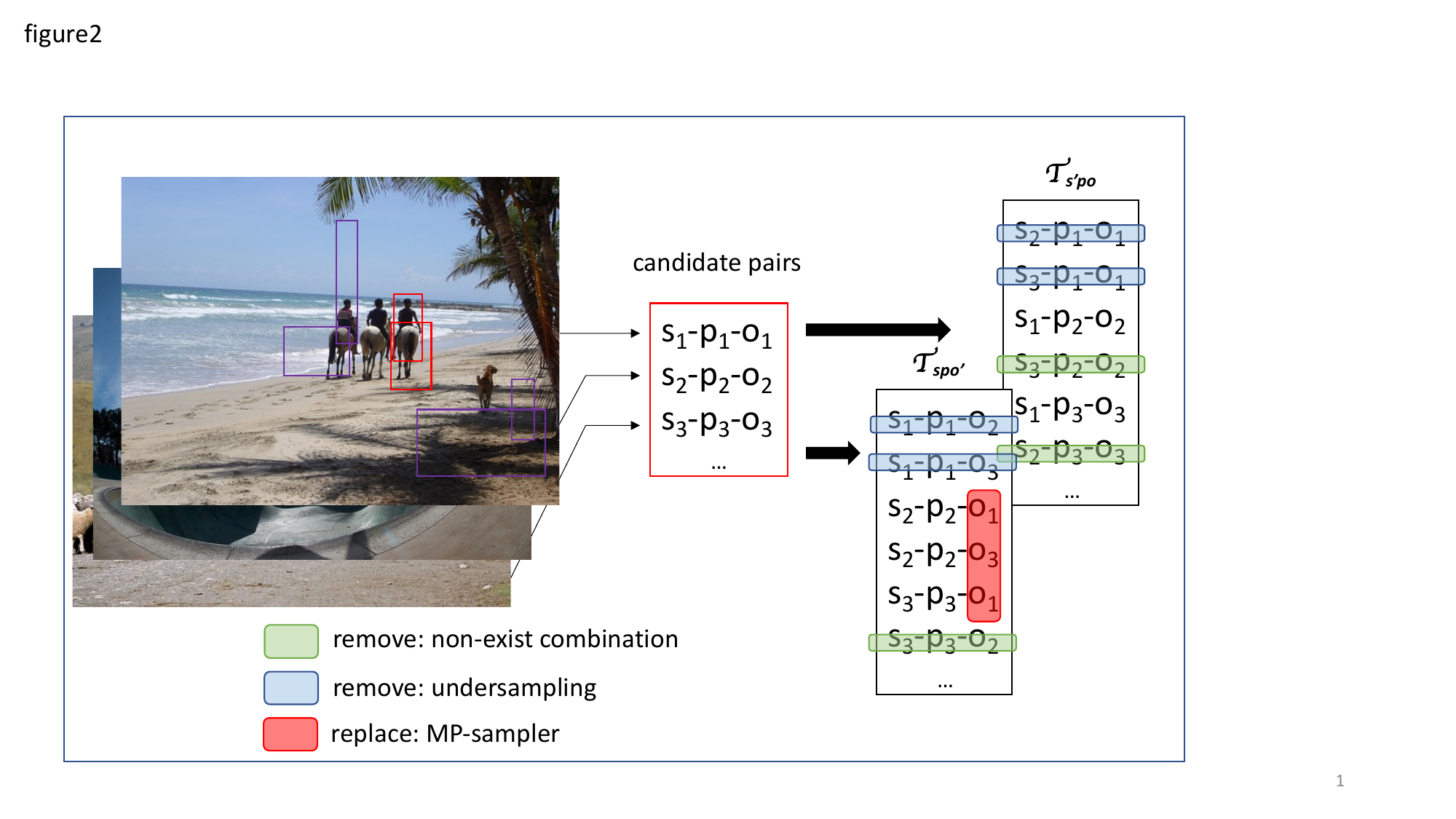}
  \caption{Building combinations from batch input proposals. Purple box pairs are excluded for low IoU with ground-truth relations and red box pairs are selected as candidates.}
  \label{fig:resampling}
\end{figure}

We present feature space triplet augmentation (FSTA) via artificial triplets. Compared with generating new image samples, features are more tractable and computationally efficient without using external knowledge \cite{SGG-NLS,heSGG_prompt_tuning}. For an input mini-batch of size $N_{B}$, the detector backbone yields RoIs of varying numbers with their object prediction results. We sample $N_{t}$ RoI pairs per image from the pool of RoI pairs that have an overlap score exceeding $s_{iou}$ with any of the ground-truth triplets in the image. Next, we enumerate a set $\mathcal{T}_{spo'}$ of combinations from these $N_{B} \times N{t}$ triplets as \verb+subject-predicate-object'+, where \verb+object'+ represents the object features from all other sampled triplets. After eliminating pairs that are absent from the training set label space, the remaining feature combinations—deemed to be reasonable—are forwarded to the same modules in the relation head. The resulting outputs are employed to compute a regularization term $\alpha\mathcal{L}_{at}$ to foster consistent predicate predictions on artificial triplets. We use the same loss function (i.e., cross entropy) as in the origin relation head for computing $\mathcal{L}_{at}$. Figure \ref{fig:resampling} visualizes our approach to building combinations for artificial triplets.

Besides generic artificial triplet synthesis, our FSTA module incorporates two novel features: (1) bi-directional resampling, and (2) Model prediction-based class sampler.

The bi-directional resampling further expands the volume of artificial triplets by enumerating \verb+subject'-predicate-object+ into a new set of combinations $\mathcal{T}_{s'po}$, where \verb+sbject'+ is sourced from other sampled triplets. As $\mathcal{T}_{s'po} \cap \mathcal{T}_{spo'} = \emptyset$, this enhances the richness and diversity of the artificial triplets. 
We define an undersampling parameter $U_{h} \in [0,1]$ to govern the predicate distribution in artificial triplets. For triplets of frequent relations (termed as ``head group''), we retain a random $U_{h}$ fraction of them. 
This can effectively achieve a distribution shift toward rare relations in artificial triplets. Overall, We combine artificial triplets built from $\mathcal{T}_{spo'}$ and $\mathcal{T}_{s'po}$.   

\begin{figure*}[t]
  \centering
  \includegraphics[trim={0.6cm 7.85cm 0.45cm 2.35cm},clip,width=\linewidth]{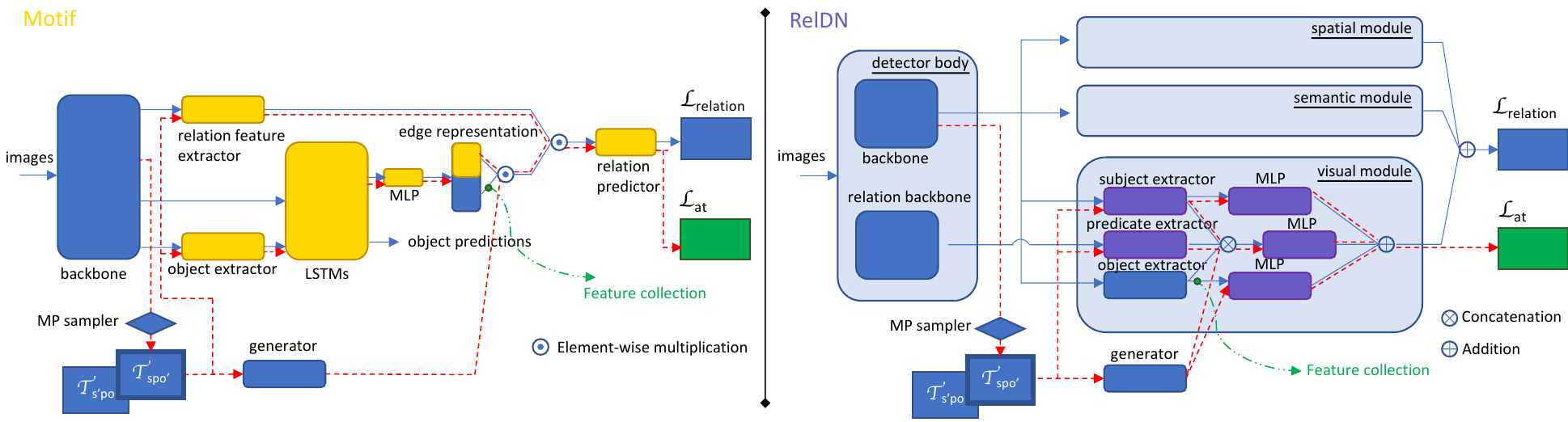}
  \caption{The schematic view illustrates the combination of FSTA and SGG models. We visualize only the flow of $\mathcal{T}_{spo'}$ with red dotted lines for readability. The green dotted line indicates the point at which features are collected in the preparation stage.}
  \label{flg:implementation}
\end{figure*}

Moreover, we propose a new sampler based on biased model predictions (hence, MP-sampler) on training data. It targets to generate suitable \verb+object'+ classes rather than mere swaps. The motivation is straightforward: \textbf{Combinations that are difficult to be predicted correctly ought to be sampled more frequently}. To begin, we enumerate the candidate \verb+object'+ classes as $\mathcal{O}_{cand}$ from the dataset label space for a given \verb+sbject-predicate+ class label pair, $(c_{s}, c_{p})$. Then, we define a difficulty score function $d(\cdot)$ which computes the mean score discrepancy between the top-1 prediction and the ground-truth predicate class: 
\begin{equation}
  \begin{aligned}
  \label{eq:mpsampler1}
  d(c_{s}, c_{p}, c_{o_{i}}) = &max(l(c_{s}, c_{p}, c_{o_{i}})) \\
  &- v(l(c_{s}, c_{p}, c_{o_{i}}), c_{o_{i}})
  \end{aligned}
\end{equation}
where $o_{i} \in \mathcal{O}_{cand}$, and $l(\cdot) \in \mathbb{R}^{\lvert \mathcal{C}_{relations} \rvert}$ returns the average post-softmax predicate prediction vector for the input combination in MP. $v(l, i)$ obtains the value of $l$ at element $i$. If the correct relation for a combination is often mispredicted, the difficulty score is positive; otherwise, it is zero. The MP-sampler then can generate an \verb+object'+ class following the probability:  
\begin{equation}
\label{eq:mpsampler2}
  p(c_{o_{i}} | (c_{s}, c_{p})) = \frac{d(c_{s}, c_{p}, c_{o_{i}})}{\sum_{j=1}^{\lvert \mathcal{O}_{cand} \rvert} d(c_{s}, c_{p}, c_{o_{j}})}  
\end{equation}
In short, our emphasis primarily rests on those hard-to-predict combinations when building artificial triplets.

With MP-sampler, we use a generator to synthesize features for \verb+object'+, as sampled classes are not assured to align with the classes from the batch's swapped features. We collect the ground-truth object features to train a conditional-GAN \cite{cGAN}. Following \cite{featGen,featgen_recon}, we define its adversarial loss function as:
\begin{equation}
\label{eq:featgen}
  \min_{G}\max_{D} \mathcal{L}_{wgangp} + \beta \mathcal{L}_{cls} + \gamma \mathcal{L}_{recon}  
\end{equation}

where $\mathcal{L}_{cls}$ and $\mathcal{L}_{recon}$ regularize the generator output via an object classifier and an reconstructor respectively, of both pre-trained on real data. Having the trained generator $G$, it is capable of yielding synthetic object features with MP-sampler, and we can construct artificial triplets for $\mathcal{T}_{spo'}$. The GAN model detail can be found in the appendix. Algorithm \ref{alg:fsta} outlines the complete procedures of FSTA.  

\begin{algorithm}[ht]
\footnotesize
\SetKwInOut{Input}{Input}
\SetKwFunction{MPsampler}{MPsampler}
\SetKwFunction{EnumValidCombination}{EnumValidCombination}
\SetKwFunction{Undersample}{Undersample}
\SetKwFunction{CrossEntropy}{CrossEntropy}
\SetKwFunction{ConCat}{ConCat}
\Input{Biased model predictions $MP$, pre-trained extracted object features and labels ${\lbrace \mathbf{f}_{t}, y_{t} \rbrace}^{T}_{t=1}$, batch size $N_{B}$, sampled pair size $N_{t}$, loss coefficient $\alpha$}

\MPsampler $\gets$ build from $MP$ based on Eq. (2) and (3);

\{${G, D}$\} $\gets$ initialize adversarial modules;

\For{$i = 1,...,k_{adv}$}{
    \{${G, D}$\} $\gets$ update as Eq. (4) using ${\lbrace \mathbf{f}_{t}, y_{t} \rbrace}$;
}

\For{$i = 1,...,k_{SGG}$}{
    \{...\};  // Do the general training step for the SGG model

    cand\_triplets $\gets$ [];
    
    \For{$j = 1,...,N_{B}$}{

        pps $\gets$ sample $N_{t}$ valid RoI pairs as shown in Fig. 3;

        cand\_triplets $\gets$ cand\_triplets + pps;
    }

    $\mathcal{T}_{spo'}$ $\gets$ \EnumValidCombination(cand\_triplets); 

    $\mathcal{T}_{s'po}$ $\gets$ \EnumValidCombination(cand\_triplets);

    $\mathcal{T}_{spo'}$ $\gets$ \Undersample($\mathcal{T}_{spo'}$);

    $\mathcal{T}_{s'po}$ $\gets$ \Undersample($\mathcal{T}_{s'po}$);
    
    gen\_obj\_labels $\gets$ \MPsampler($\mathcal{T}_{spo'}$);
    
    gen\_obj\_features $\gets$ $G$($\mathbf{z}$, gen\_obj\_labels);
    
    $\mathcal{T}_{spo'}$ $\gets$ replace object part with \{gen\_obj\_labels, gen\_obj\_features\};

    $\mathcal{L}_{at}$ $\gets$ \CrossEntropy($M$($F$(\ConCat($\mathcal{T}_{spo'}$, $\mathcal{T}_{s'po}$))), ground\_truth\_relations);
    
    update model from $\alpha$ $\mathcal{L}_{at}$;
}  
\caption{FSTA}
\label{alg:fsta}
\end{algorithm}

\subsection{Soft Transfer}
\label{softlabel}

The IETrans internal transfer reassigns relation labels from the general (source) ones to the informative (target) ones. However, some transfers are suboptimal: human evaluation deems only 76\% of general-informative pairs as ``reliable'' \cite{IETrans}. While tail performance can benefit from these transfers, the cost of head performance drop is a concern.

A finer control on individual transfers could improve the transfer efficiency and thus alleviate the tail-head performance trade-off. Instead of a complete label transfer from a general ($p\rightarrow0$) to an informative predicate ($p\rightarrow1$), we propose the Soft Transfer that assigns non-binary probabilities to source and target predicate classes. Soft Transfer consists of two steps. Firstly, we rank all the reassigned pairs using a triplet-wise reliability score, from which pairs are selected for Soft Transfer. Second, a mapping function converts the reliability score into probabilities for source and target labels.  

Based on the observation that transfer reliability varies from one combination to another, we define a preliminary function $r_{int}(\cdot)$ to estimate the degree of reliability. Given an transfer decision list, each list item includes a triplet index $i$, a source class $c_{p}^{src,i}$, and a target class $c_{p}^{tar,i}$. To determine the reliability score, we use the prediction output difference following 
\begin{equation}
\label{eq:reliability score}
  r_{int}(i) = v(l_{triplet}(i), c_{p}^{tar,i}) -v(l_{triplet}(i), c_{p}^{src,i})  
\end{equation}
where $l_{triplet}(i) \in \mathbb{R}^{\lvert \mathcal{C}_{relations} \rvert}$ returns the post-softmax model prediction of triplet $i$, and $v(l_{triplet}(\cdot), j)$ retrieves the value of $l_{triplet}(\cdot)$ at class $j$. We rank the score in ascending order and pick the top $k_{s}$\% triplets for Soft Transfer while the other remains. 

For those triplets with low reliability scores, we consider them as over-transferred. Thus, a positive probability should be assigned to the source class as the ground-truth label instead of zero. 
While achieving this and ensuring that the sum of the classes for the label is $1$, we map the reliability scores to values within the range $[0,1]$. Given a mapping function $Q(\cdot)$, the post-transferred result for triplet $i$ can be represented as its label probability:

\begin{equation}
    label_{i}(c) = 
    \begin{cases}
        \frac{1}{1+Q(r_{int}(i))},& \text{if } c = c_{p}^{tar,i}\\
        \frac{Q(r_{int}(i))}{1+Q(r_{int}(i))},& \text{if } c = c_{p}^{src,i}\\
        0,              & \text{otherwise}
    \end{cases}
\end{equation}
we set $Q(\cdot)=1-Q'(\cdot)$, where $Q'(\cdot)$ is a linear min-max scaling for the reliability scores. 

Soft Transfer is applied on the original relation loss and needs no changes to it. Table \ref{table:ST_example} shows an example of post-transferred annotations. In IETrans, predicates of the selected triplets are reassigned to more informative ones (red). Our Soft Transfer evaluates the reliability of these reassigned predicates and converts them to non-binary values (blue).

\begin{figure}[h]
  \centering
  \includegraphics[width=0.7\linewidth]{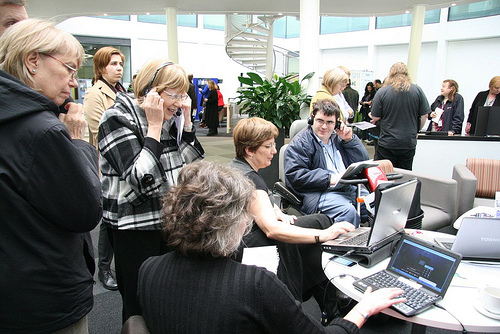}
  \caption{An example training image in VisualGenome.}
  \label{fig:enh dataset example}
\end{figure}
\vspace{-5mm}
\begin{table}[h]
    \centering
    \footnotesize
    \caption{The differences in relation annotation among the raw dataset, the baseline IETrans (excluding External Transfer), and our proposed Soft Transfer for Figure \ref{fig:enh dataset example}.}
    \begin{tabular}{lll}
        \hline
        config & \multicolumn{2}{l}{relation annotations}\\
        \hline
        raw & man-sitting on-chair & laptop-on-table\\
        & plant-in-pot & person-on-laptop\\
        \hline
        +IETrans  & man-sitting on-chair & laptop-\textcolor{red}{above}-table\\
        & plant-in-pot & person-\textcolor{red}{looking at}-laptop\\
        \hline
        ++SoftTrans & man-sitting on-chair & \\
        (ours) & \multicolumn{2}{l}{laptop-\textbf{\textcolor{blue}{(above:0.76, on:0.24)}}-table} \\
        & plant-in-pot & \\
        & \multicolumn{2}{l}{person-\textbf{\textcolor{blue}{(looking at:1.0, on:0.0)}}-laptop} \\
        \hline
    \end{tabular}
    \label{table:ST_example}
\end{table}

\begin{table*}[t]
  \caption{The performance comparison for the predcls task on VG150. Scores for models listed in the first section are cited from their original papers, while models in subsequent sections use our implementation. ``Model++X'' is shorthand for ``Model+IETrans+X''. The best overall scores within each section are highlighted in bold. (Unit: \%) 
  }
  \footnotesize
  \label{tab:predcls}
  \begin{tabular}{lcccccccc}
    \toprule
           &  \multicolumn{8}{c}{Predicate Classification (Predcls)} \\
    \cmidrule(lr){2-9}
    models &R@50 & R@100 & mR@50(h/b/t) & mR@100(h/b/t) & F1@50 & F1@100 & A@50 & A@100\\
    \midrule
    Motif+TDE \cite{TDE} & 46.2 & 51.4 & 25.5(-) & 29.1(-) & 32.9 & 37.2 & 35.9 & 40.3\\
    Motif+DLFE \cite{DLFE} & 52.5 & 54.2 & 26.9(-) & 28.8(-) & 35.6 & 37.6 & 39.7 & 41.5\\
    Motif+NICE \cite{NICE}  & 55.1 & 57.2 & 29.9(-) & 32.3(-) & 38.8 & 41.3 & 42.5 & 44.8 \\
    Motif+IETrans \cite{IETrans} & 54.7 & 56.7 & 30.9(-) & 33.6(-) & 39.5 & 42.2 & 42.8 & 45.2 \\
    Motif+IETrans+rwt \cite{IETrans} & 48.6 & 50.5 & 35.8(-) & 39.1(-) & 41.2 & 44.1 & 42.2 & 44.8 \\
    Motif+Inf \cite{inf_CVPR23} & 51.5 & 55.1 & 24.7(-) & 30.7(-) & 33.4 & 39.4 & 38.1 & 42.9 \\
    \midrule
    Motif                  & 65.0 & 67.2 & 16.1(39.3/9.3/1.2)   & 17.8(42.3/11.1/1.3)  & 25.8 & 28.1 & 40.6 & 42.5\\
    Motif+IETrans\dag      & 54.8 & 57.1 & 29.6(42.2/33.4/14.0) & 32.9(45.8/37.2/16.4) & 38.4 & 41.7 & 42.2 & 45.0\\
    Motif++FSTA (ours)      & 54.0 & 56.2 & 31.0(42.4/33.2/18.2) & 34.8(45.8/37.4/22.0) & \textbf{39.4} & \textbf{43.0} & 42.5 & 45.5\\
    Motif++SoftTrans (ours) & 58.6 & 60.8 & 28.0(42.5/31.8/10.6) & 30.8(46.0/35.1/12.3) & 37.9 & 40.9 & 43.3 & 45.8\\
    Motif++Full (ours)       & 57.1 & 59.4 & 29.8(41.6/32.0/16.5) & 33.2(45.1/35.8/19.5) & 39.2 & 42.6 & \textbf{43.5} & \textbf{46.3}\\
    \midrule
    Motif+IETrans+rwt\dag & 51.5 & 53.7 & 34.4(43.2/37.3/23.4) & 38.8(46.3/40.5/30.2) & 41.3 & 45.1 & 43.0 & 46.2 \\
    Motif++FSTA+rwt (ours) & 49.0 & 51.1 & 35.9(42.0/36.8/29.1) & 40.6(45.1/40.1/36.8) & 41.4 & 45.2 & 42.4 & 45.8 \\
    Motif++SoftTrans+rwt (ours) & 55.6 & 57.8 & 33.1(43.3/36.1/20.5) & 38.3(46.4/39.9/28.9) & 41.5 & 46.0 & \textbf{44.4} & \textbf{48.0} \\
    Motif++Full+rwt (ours) & 53.4 & 55.5 & 34.7(42.4/35.7/26.6) & 39.5(45.4/39.1/34.2) & \textbf{42.1} & \textbf{46.1} & 44.1 & 47.5 \\
    \midrule
    \midrule
    RelDN                  & 60.7 & 62.2 & 13.8(38.5/4.2/0.0) & 14.9(40.9/5.3/0.1) & 22.5 & 24.0 & 37.3 & 38.6\\
    RelDN+IETrans\dag      & 38.6 & 39.9 & 29.6(35.4/33.1/20.5) & 32.3(37.7/36.2/23.3) & 33.5 & 35.7 & 34.1 & 36.1\\
    RelDN++FSTA (ours)      & 37.0 & 38.2 & 31.3(33.7/33.0/27.3) & 34.1(35.8/36.0/30.5) & 33.9 & 36.0 & 34.2 & 36.2\\
    RelDN++SoftTrans (ours) & 52.4 & 53.8 & 26.1(37.6/27.6/13.9) & 28.1(40.0/29.7/15.3) & 34.8 & 36.9 & \textbf{39.3} & \textbf{41.0}\\ 
    RelDN++Full (ours)  & 49.2 & 50.7 & 28.2(35.8/28.2/21.0) & 30.6(37.9/30.8/23.5) & \textbf{35.9} & 
    \textbf{38.2} & 38.7 & 40.7\\ 
    \midrule
    RelDN+IETrans+rwt\dag & 25.2 & 26.3 & 32.1(28.7/35.5/32.0) & 34.6(30.7/37.8/35.1) & 28.2 & 29.9 & 28.7 & 30.5 \\
    RelDN++FSTA+rwt (ours)& 24.2 & 25.3 & 32.5(28.2/35.6/33.5) & 35.8(30.1/38.1/38.8) & 27.7 & 29.6 & 28.4 & 30.5 \\
    RelDN++SoftTrans+rwt (ours)& 36.1 & 37.4 & 31.6(34.7/32.7/27.8) & 34.8(36.8/34.9/32.7) & \textbf{33.7} & \textbf{36.1} & \textbf{33.9} & \textbf{36.1} \\
    RelDN++Full+rwt (ours) & 33.6 & 34.9 & 31.7(32.6/32.3/30.2) & 35.1(34.6/35.3/35.4) & 32.6 & 35.0 & 32.7 & 35.0 \\
    
    \bottomrule
  \end{tabular}
\end{table*}

\subsection{Implementation Details}
\label{3_5_implementation}
We build our work upon an open source SGG model implementation\footnote{\footnotesize \url{https://github.com/microsoft/scene_graph_benchmark}\label{fn1}} \cite{SGGBench}. We integrate our system into two prevalent SGG model of distinct types: Motif \cite{Motif} (which employs LSTM) and RelDN \cite{RelDN} (which utilizes CNN, multi-modality fusion, and contrastive losses). These were selected because they represent a variety of design elements commonly found in popular models. We use a ResNet50-FPN \cite{ResNet} Faster-RCNN \cite{FasterRCNN} as the common detector backbone. The detector backbone is pre-trained on VisualGenome \cite{VisualGenome} and kept frozen. Fig. \ref{flg:implementation} illustrates how to combine our module with these SGG models. We implement IETrans with the default parameters: $k_{i}=70$ and $k_{e}=100$. 
In the FSTA module, $N_{t}$ is set to 2 for Motif and 5 for RelDN to balance the number of artificial triplets in a mini-batch, considering the smaller batch size for RelDN. We set $s_{iou}=0.7$\footnote{We follow the implementations in \ref{fn1} to compute $s_{iou}$.}, $U_{h}=0.2$ for Motif, and $s_{iou}=0.5$, $U_{h}=0.8$ for RelDN. Both models have a loss coefficient of $\alpha=0.1$. We omit artificial triplets from $\mathcal{T}_{s'po}$ if their predicates are not in the tail group. For the soft transfer module, we set $k_{s}$ to 10 for Motif and 70 for RelDN. In the experiments with a ``reweighting'' setting, only the original loss function is applied with reweighting, while $\mathcal{L}_{at}$, the loss for FSTA, does not apply as a standalone module. $k_{s}$ also changes to 30 for Motif and 90 for RelDN.

\section{Experiments}
\label{sec: exp}

\subsection{Dataset and Evaluation Protocol}
\label{sec: dataset}

We evaluated our system on the benchmark VG150 split of the VisualGenome dataset. This dataset consists of 60,784 training images and 26,446 testing images. It contains 150 object classes and 50 relation classes. Following the approach of \cite{DC2-ABCS}, we sorted the predicates by cardinality, grouping the top 16, middle 17, and bottom 17 into \textbf{head}, \textbf{body}, and \textbf{tail} groups, respectively. 

\begin{table*}[t]
  \caption{The performance comparison for the sgcls and sgdet task on VG150. ``Model++X'' is shorthand for ``Model+IETrans+X''. Best overall scores in the section are highlighted in bold. Full results can be found in the appendix. (Unit: \%)}
  \label{tab:sgcls+sgdet}
  \footnotesize
  \begin{tabular}{lcccccccc}
    \toprule
           &  \multicolumn{4}{c}{Scene Graph Classification (Sgcls)} & \multicolumn{4}{c}{Scene Graph Detection (Sgdet)} \\
    \cmidrule(lr){2-5}\cmidrule(lr){6-9}
    models & R@100 & mR@100(h/b/t) & F1@100 & A@100 & R@100 & mR@100(h/b/t) & F1@100 & A@100\\
    \midrule
    Motif                  & 38.9 & 10.7(25.2/7.0/0.8) & 16.8 & 24.8 & 37.7 & 9.3(23.0/5.6/0.2) & 14.9 & 23.5 \\
    Motif+IETrans\dag      & 30.1 & 20.9(25.9/21.2/15.9) & \textbf{24.7} & 25.5 & 29.2 & 16.5(24.9/18.4/6.8) & 21.1 & 22.9\\
    Motif++FSTA (ours)     & 30.5 & 20.6(26.0/21.0/15.1) & 24.6 & 25.6 & 28.8 & 17.1(25.1/18.3/8.2) & 21.5 & 23.0 \\
    Motif++SoftTrans (ours)& 34.1 & 18.7(26.4/20.8/9.3) & 24.2 & \textbf{26.4} & 32.2 & 15.8(25.2/18.6/4.1) & 21.2 & 24.0\\
    Motif++Full (ours)     & 33.3 & 19.2(25.7/20.5/11.8) & 24.4 & 26.3 & 32.2 & 17.0(24.6/18.5/8.3) & \textbf{22.3} & \textbf{24.6} \\
    \midrule
    \midrule
    RelDN                  & 36.9 & 7.9(22.9/1.6/0.0) & 13.0 & 22.4 & 38.0 & 8.2(23.7/1.9/0.0) & 13.5 & 23.1\\
    RelDN+IETrans\dag      & 23.3 & 19.0(22.0/21.2/14.1) & 20.9 & 21.2 & 22.0 & 18.4(22.0/20.9/12.4) & 20.0 & 20.2\\
    RelDN++FSTA (ours)     & 22.8 & 19.3(21.6/21.2/15.2) & 20.9 & 21.1 & 21.1 & 19.1(21.3/21.0/15.1) & 20.1 & 20.1\\
    RelDN++SoftTrans (ours)& 32.8 & 15.5(23.4/16.9/6.7) & 21.1 & \textbf{24.2} & 32.3 & 15.4(23.8/16.5/6.3) & 20.9 & \textbf{23.9}\\
    RelDN++Full (ours)     & 31.1 & 16.8(22.9/17.4/10.6) & \textbf{21.8} & 24.0 & 29.3 & 17.2(22.5/17.8/11.5) & \textbf{21.7} & 23.3\\
    \bottomrule
  \end{tabular}
\end{table*}

Our analysis focused on standard SGG tasks: \textbf{predcls}, \textbf{sgcls}, and \textbf{sgdet} \cite{Motif,VCTree,RelDN}. These tasks evaluate the model with incrementally higher demands. For instance, ``predcls'' only assesses the model's ability in classifying relations using given object locations and categories. In contrast, ``sgdet'' evaluates both relation classification and object detection simultaneously. Our primary attention was on predcls since our proposed modules target improving predicate classification performance. We used the \textbf{Recall(R)@K} and \textbf{mean Recall(mR)@K} metrics for both full test set and per-class averaged recall evaluations. It is noteworthy that Recall@K is dominated by the performance of the top frequent classes due to the skewed predicate distribution, whereas mean Recall@K treats all classes equally. Given the observed trade-off between Recall@K and mean Recall@K from earlier studies, we also reported the \textbf{F1@K} (their harmonic mean) and \textbf{Avg(A)@K} (their arithmetic mean) \cite{IETrans,NICE} as the ``overall'' metrics in our comprehensive evaluation. \textbf{All metrics are the higher the better}.

\subsection{Comparing to Other Methods}
 
We compared our results with IETrans and several other recent model-agnostic SGG methods (first section of Table \ref{tab:predcls}). 
IETrans serves as the baseline and is currently one of the best model-agnostic methods available. 

\noindent{\textbf{Original baseline and our re-implementation.}} 
We compared our method with the reproduced baselines (denoted as \dag). For Motif+IETrans in the predcls task, the reproduced version yielded similar scores to those of the original, with a slightly higher R@100 and lower mR@100. These differences may due to some implementation variations in the base SGG model. Therefore, we use the reproduced version as our standard because it maintains identical implementation settings and IETrans transfer lists, consistent with our proposed methods. The original IETrans paper did not present results for RelDN; therefore, we also compared our results with a reproduced version. In summary, all our implementations share the basic settings to ensure a fair comparison.

\noindent{\textbf{Improved relation prediction over the baseline.}}
Table \ref{tab:predcls} summarizes the scores for predcls. The results reveal that our method substantially outperformed the baseline for both the Motif and RelDN models. Specifically, the F1@100 score rose from 41.7 to 43.0 (a 3.1\% relative gain) for Motif, and from 35.7 to 38.2 (a 7.0\% relative gain) for the RelDN model. The A@100 score also increases. With FSTA, the standout feature was the mR enhancement in tail classes (e.g., from 16.4 to 22.0 for Motif). The artificial triplets generated in FSTA enriched the variation of triplets available for the relation predictor, aiding especially the sparse classes. As for frequent classes, the score decline is minor.
On the other hand, Soft Transfer was intended to reduce the degree of label reassignment for less reliable transfers. This led to a score trend the opposite of the original IETrans: while recall scores raised, the tail mean recall scores decreased (e.g., R@100 increases from 57.1 to 60.8 for Motif, and 39.9 to 53.8 for RelDN). In certain cases, Soft Transfer can slightly reduce the F1 score, because the harmonic mean prioritizes enhancements in the smaller one. Nonetheless, the Avg@100 witnessed a notable boost with Soft Transfer. Combining both modules, the full system leveraged their complementary benefits, consistently delivering among the top F1/Avg@100 results for both models, indicating an effective balance of trade-offs.

\noindent{\textbf{Compatible with the reweighting setting.}}
We also adhere to the original settings described in the IETrans paper to compare the models when integrated with the ``reweighting'' technique (+rwt) \cite{IETrans}. Our method proved efficacious even under this setting. Both the FSTA and Soft Transfer modules served their intended purposes, driving improvements across rare and frequent classes alike. The ``Motif++Full+rwt'' method achieves an increase in F1@100 from 45.1 to 46.1, and in Avg@100 from 46.2 to 47.5, thereby demonstrating the mitigation of the performance trade-off. For RelDN with reweighting, the performance was not as beneficial as for the Motif models. Although the mR@100 for the tail group further increased, the impact on R@K cannot be overlooked, leading to a dip in the overall scores. One possible explanation for this could be the architecture of the RelDN model, which already incorporates a frequency prior branch. Consequently, we did not add the frequency prior values during inference for the RelDN models, while we did so for the Motif models following \cite{IETrans}. This leads to a more serious degradation in head classes, despite having the best performance on tail classes. However, our method still consistently surpassed the baseline in the RelDN +rwt setting.

\noindent{\textbf{Similar trends observed for sgcls and sgdet.}}  
Table \ref{tab:sgcls+sgdet} showcases the digested results for sgcls and sgdet. Here, we noticed trends analogous to those in predcls. For RelDN, the full version achieves the best F1@100 and the second-best A@100. For the Motif sgdet, the full version outperforms all the others on both overall metrics (i.e., a 5.7\% and 7.4\% relative gain for F1@100 and A@100 over the baseline method, respectively). However, the FSTA module yields some unexpected results in the sgcls task. One potential cause is that we applied identical FSTA settings across all tasks. However, sgcls uniquely relies on ground-truth boxes only for input proposals, which is different from the other tasks. This difference might result in distinct regularization effects, as the artificial triplets are constructed from sampled proposal pairs.

\section{Discussions}

In this section, we focus on the predcls task results for the Motif model\footnote{Unless specified otherwise, this means ``Model+IETrans+X''.} to gain a deeper understanding of our methods. Results for RelDN can be found in appendix. 

\subsection{Ablation Study}
Table~\ref{tab:fsta ablation} presents the components ablated from FSTA to demonstrate their contributions to the module. The results indicate that all components positively influence the improvement of F1@100. Among these, undersampling has the greatest impact on F1@100, adjusting the proportions of artificial triplets in the head, body, and tail predicate groups from 0.70, 0.14, and 0.15 to 0.33, 0.32, and 0.35 respectively. Additionally, incorporating $\mathcal{T}_{s'po}$ effectively introduces new training combinations. The MP-sampler also plays a crucial role, further boosting R@100.

Table~\ref{tab:abl-motif-rwt} summarizes the ablation results with reweighting. A similar trend is observed that the components in FSTA contribute to the increase in scores for tail relation groups and the mR@100.

\vspace{-5mm}
\begin{table}[h]
  \caption{The ablation study results for our FSTA module. For the components, "us" refers to undersampling, "+sbj" denotes adding artificial set $\mathcal{T}_{s'po}$, and MP indicates that the MP-sampler is applied. (Unit: \%)}
  \label{tab:fsta ablation}
  \footnotesize
  \begin{tabular}{cccccc}
    \toprule
    
    us   & +sbj       & MP         & R@100 & mR@100(h/b/t) & F1/Avg@100\\
    \midrule
               &            &            & 55.8 & 33.4(45.2/38.7/17.1) & 41.8/44.6 \\
    \checkmark &            &            & 55.3 & 34.4(45.3/39.9/19.6) & 42.4/44.9 \\
    \checkmark & \checkmark &            & 54.8 & 35.0(44.8/39.1/21.6) & 42.7/44.9 \\
    \checkmark & \checkmark & \checkmark & 56.2 & 34.8(45.8/37.4/22.0) & \textbf{43.0}/\textbf{45.5} \\
    \bottomrule
  \end{tabular}
\end{table}
\vspace{-10mm}
\begin{table}[h]
  \caption{The ablation study results for our FSTA module with Motif and ``reweighting''. Item descriptions are identical to Table 4. (Unit: \%)}
  \label{tab:abl-motif-rwt}
  \footnotesize
  \begin{tabular}{cccccc}
    \toprule
    
    us   & +sbj       & MP         & R@100 & mR@100(h/b/t) & F1/Avg@100\\
    \midrule
               &            &            & 54.7 & 36.6(45.8/39.2/25.4) & 43.9/45.7 \\
    \checkmark &            &            & 53.1 & 38.7(45.1/40.8/30.7) & 44.8/45.9 \\
    \checkmark & \checkmark &            & 52.5 & 39.7(44.5/41.4/33.4) & 45.2/\textbf{46.1} \\
    \checkmark & \checkmark & \checkmark & 51.1 & 40.6(45.1/40.1/36.8) & \textbf{45.2}/45.8 \\
    \bottomrule
  \end{tabular}
\end{table}
\vspace{-5mm}

\subsection{Sensitivity Analysis}

We then investigate the choice of percentage $k_{s}$ in Soft Transfer. 
A ``Na\"ive'' setting would simply apply soft transfer to all reassigned triplets without our ranking and mapping mechanisms, where both source and target labels are assigned a value of 0.5. The results are summarized in Table \ref{tab:soft transfer sensitivity}. 

As the number of entirely transferred triplets is reduced, R@100 recovers as $k_{s}$ grows, yet mR@100 decreases. The ``Na\"ive'' setting consistently performs worse than the others in the F1@100 or even Avg@100 metrics and only be on par with the baseline IETrans ($Avg@100=45.0$). This highlights the significance of our devised method. 

\vspace{-5mm}
\begin{table}[h]
  \caption{The sensitivity results for our Soft Transfer module. ``$\ast$'' indicates the setting applied in our method. (Unit: \%)}
  \footnotesize
  \label{tab:soft transfer sensitivity}
  \begin{tabular}{lccc}
    \toprule
    
    settings:         & R@100 & mR@100(h/b/t)        & F1/Avg@100 \\
    \midrule
    ${k_{s}=0.1}\ast$ & 60.8  & 30.8(46.0/35.1/12.3) & \textbf{40.9}/45.8 \\
    $k_{s}=0.3$       & 63.1  & 30.1(45.4/34.5/11.3) & 40.7/\textbf{46.6} \\
    $k_{s}=0.5$       & 64.5  & 28.4(44.7/31.6/9.8)  & 39.4/46.4 \\
    Na\"ive           & 66.5  & 23.4(43.6/22.6/5.0)  & 34.6/45.0 \\
    \bottomrule
  \end{tabular}
\end{table}
\vspace{-5mm}

\subsection{Comparison with Real Data Resampling}

We compare the effects of resampling real data with our FSTA. Although both approaches augment the number of rare predicates, their motivations and methods differ. FSTA aims to steer the predicate classification layers towards understandind the inherent concept of the predicate by leveraging combinative, yet semantically plausible, artificial triplets. This also introduces new variations to the training data. In contrast, resampling merely duplicates samples from rare classes to mitigate dataset imbalance; however, it is susceptible to overfitting.

For our real data resampling implementation, we altered the training set by duplicating the image $n$ times if it contained more than one triplet of tail group predicates. 
We analyzed the performance difference when applied independently and the combined for Motif+IETrans on the predcls task. The scores are listed in Table \ref{tab:FSTA resampling}. 

Our "+FSTA" is more effective than "+resampling", as it yields superior overall metrics for both F1 and Avg. Combining both can boost the mean recall of the tail group. Intensive resampling improved tail classes but reduced frequent class recall. We did not observe positive effects when n was larger than 4.

\vspace{-5mm}
\begin{table}[h]
  \caption{The results of comparing our FSTA with resampling. (Unit: \%)}
  \footnotesize
  \label{tab:FSTA resampling}
  \begin{tabular}{lccc}
    \toprule
    
    settings:         & R@100 & mR@100(h/b/t) & F1/Avg@100 \\
    \midrule
    IETrans   & 57.1  & 32.9(45.8/37.2/16.4) & 41.7/45.0 \\
    \midrule
    +resample($n=1$)   & 57.2  & 33.5(45.5/37.8/18.1) & 42.3/45.4 \\
    +resample($n=2$)   & 55.6  & 33.5(45.1/37.4/18.8) & 41.8/44.5 \\
    +resample($n=3$)   & 55.5  & 33.5(44.6/36.8/19.9) & 41.8/44.5 \\
    \midrule
    +FSTA (ours)   & 56.2  & 34.8(45.8/37.4/22.0) & 43.0/\textbf{45.5} \\
    +resample+FSTA   & 54.9  & 35.8(44.7/37.2/26.1) & \textbf{43.3}/45.3 \\
    \bottomrule
  \end{tabular}
\end{table}
\vspace{-5mm}

\subsection{Parameter Choices for FSTA}

To explore the quality of our $s_{iou}$ and $U_h$ choices, which are applied across settings, we assess the performance within the reweighting setting. Table~\ref{tab:sensi-motif-rwt} details the results.

Our observations are as follows: (1) Lower values of $U_h$ tend to result in higher tail group performance, attribute to the increased ratio of tail relations in the artificial triplets. (2) A higher $s_{iou}$ threshold allows only the most precise feature representations, benefiting the data-sparse tail group while potentially harming the generalizability in frequent classes.
Overall, we found that the parameters we selected perform reasonably well, even under such a different setting.

\vspace{-5mm}
\begin{table}[h]
  \caption{The results of parameter choices for FSTA with Motif and ``reweighting''. ``$\ast$'' indicates the setting applied in our method. (Unit: \%)}
  \footnotesize
  \label{tab:sensi-motif-rwt}
  \begin{tabular}{ll|ccc}
    \toprule
    
    Param. & Value    & R@100 & mR@100(h/b/t)        & F1/Avg@100 \\
    \midrule
    \multirow{5}{*}{$s_{iou}$} & 0.5 & 51.6 & 39.9(45.8/40.6/33.9) & \underline{45.0}/\underline{45.8} \\ 
    &0.6           & 50.9 & 39.2(45.8/41.0/31.2) & 44.3/45.1 \\
    &0.7$\ast$       & 51.1 & 40.6(45.1/40.1/36.8) & \textbf{45.2}/\textbf{45.9} \\ 
    &0.8           & 50.3 & 40.7(44.7/40.0/37.6) & 45.0/45.5 \\
    &0.9           & 49.5 & 40.9(44.4/39.4/39.1) & 44.8/45.2 \\
    \midrule
    \multirow{5}{*}{$U_h$} &0.2$\ast$ & 51.1 & 40.6(45.1/40.1/36.8) & \underline{45.2}/\underline{45.9} \\ 
    &0.4           & 51.8 & 40.3(45.4/40.5/35.4) & \textbf{45.3}/\textbf{46.1} \\
    &0.5           & 51.0 & 40.2(45.3/40.5/35.1) & 45.0/45.6 \\ 
    &0.6           & 51.2 & 39.7(45.8/39.5/34.3) & 44.7/45.5 \\
    &0.8           & 51.7 & 39.9(44.5/43.1/32.3) & 45.0/45.8 \\
    \bottomrule
  \end{tabular}
\end{table}

\vspace{-5mm}

\subsection{A Study on MP-sampler}
 
\begin{figure}[h]
    \centering
    \begin{subfigure}{0.48\columnwidth}
        \includegraphics[width=\linewidth]{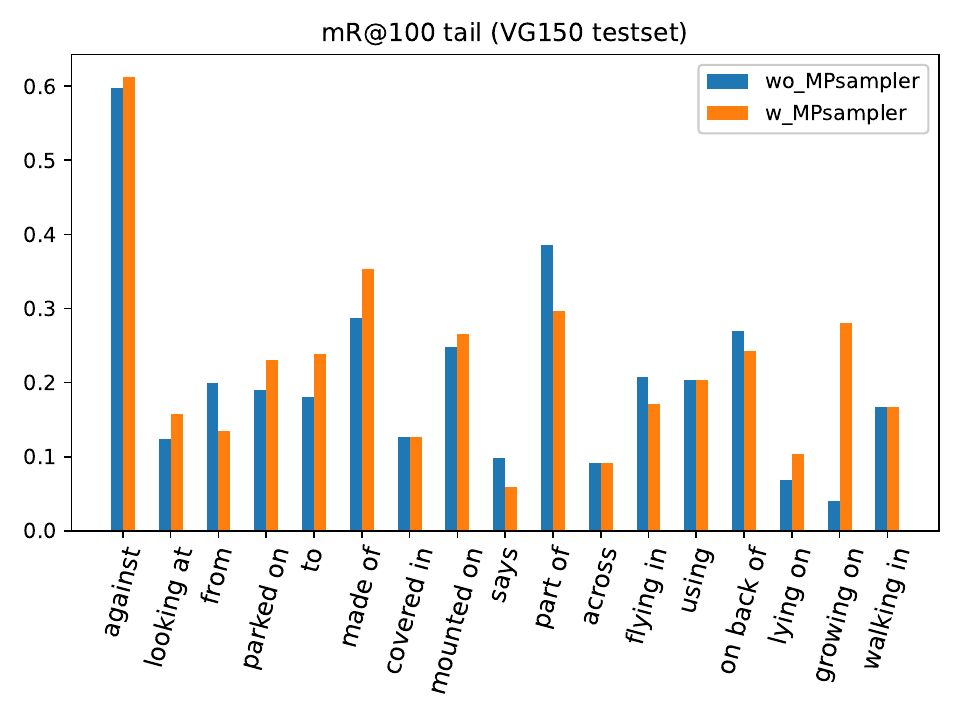}
    \end{subfigure}
    \begin{subfigure}{0.48\columnwidth}
        \includegraphics[width=\linewidth]{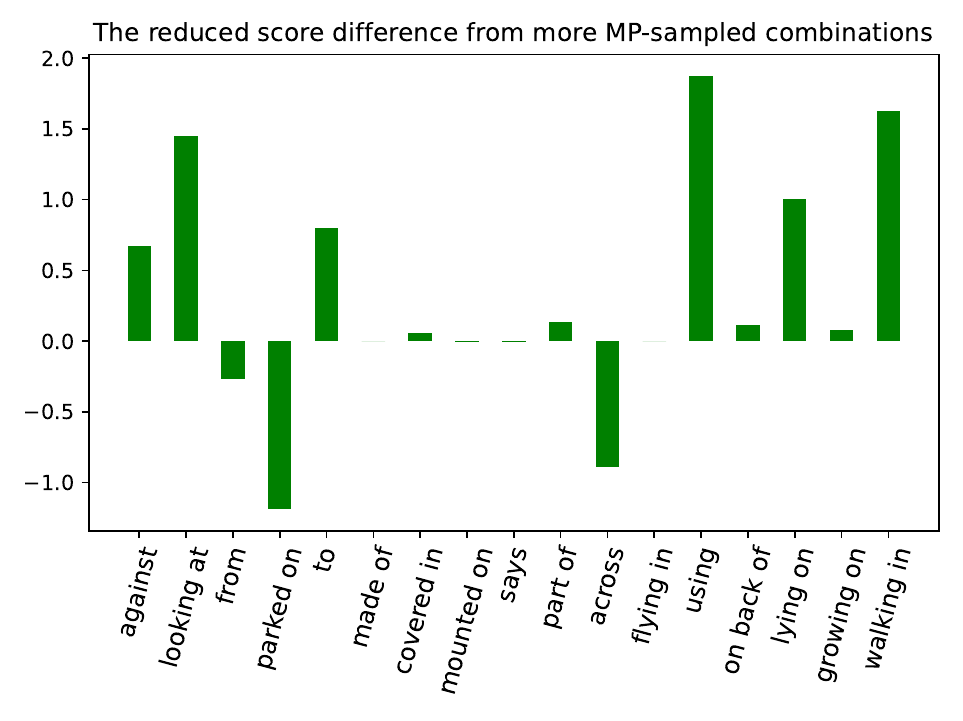}
    \end{subfigure}
    \caption{A case study examining the effects of the MP-sampler. (Left) The tail group mR@100 comparison between setups without and with MP-sampler. (Right) The accumulated $d(\cdot)$ reduction contributed by objects that are sampled more frequently.}
    \label{fig:mp sampler}
\end{figure}

We examine the role of the MP-sampler. 
In this case, the count of artificial triplets per predicate class is invariant, but the distribution of combinations changes. We undertake a case study focusing on the mR@100 tail group, where FSTA has shown significant performance gains. Fig. \ref{fig:mp sampler} (left) portrays the per-class recall: 12 of 17 classes either tie or improve with the MP-sampler. Next, we study the rationale behind the signal designed in the MP-sampler. It uses the scores from $d(\cdot)$ as the sampling probability, which inversely correlates with recall. For example, $d(\cdot)=0$ implies that the top-1 predicate prediction aligns with the ground-truth, whereas $d(\cdot)>0$ does not. We hypothesize that sampling more object labels of higher $d(\cdot)$ scores can make correct predictions easier for the unbised model.  
Thus, we analyze the changes in accumulated $d(\cdot)$ scores between the pretrained and unbiased models, from those object classes where counts have increased.
Fig. \ref{fig:mp sampler} (right) visualizes the results computed on test data. 
Out of the 17 classes, 14 show non-negative total reductions scores, confirming that our designed signal is evident in the test data.
We also inspect the relationship between the reduced score and class recall. The majority of classes follow a similar trend, with only four exhibiting the converse pattern (e.g., recall increases while the reduced score is negative). 

\subsection{Feature Visualization}
\begin{figure}[ht]
    
    \centering
    \begin{subfigure}{0.48\columnwidth}
        \includegraphics[width=\linewidth]{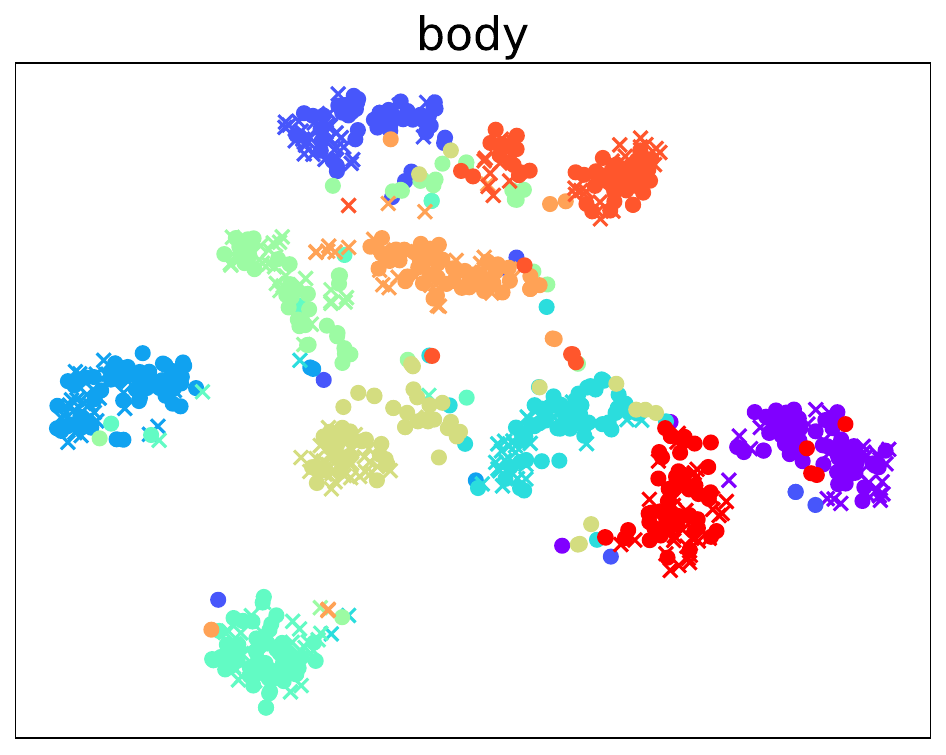}
    \end{subfigure}
    \begin{subfigure}{0.48\columnwidth}
        \includegraphics[width=\linewidth]{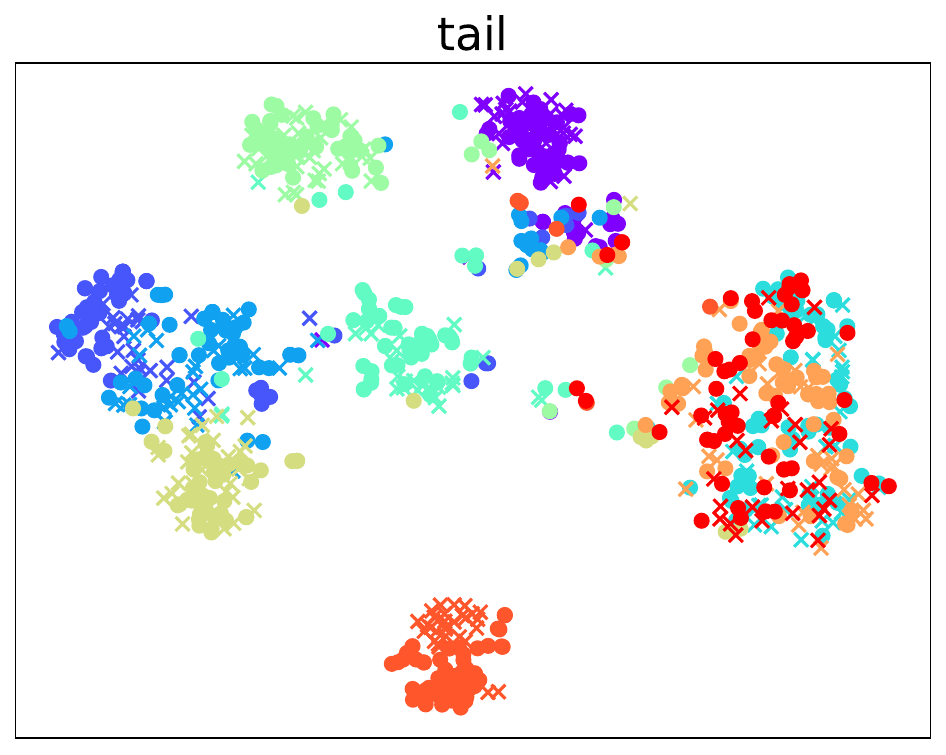}
    \end{subfigure}
    \begin{subfigure}{0.48\columnwidth}
        \includegraphics[width=\linewidth]{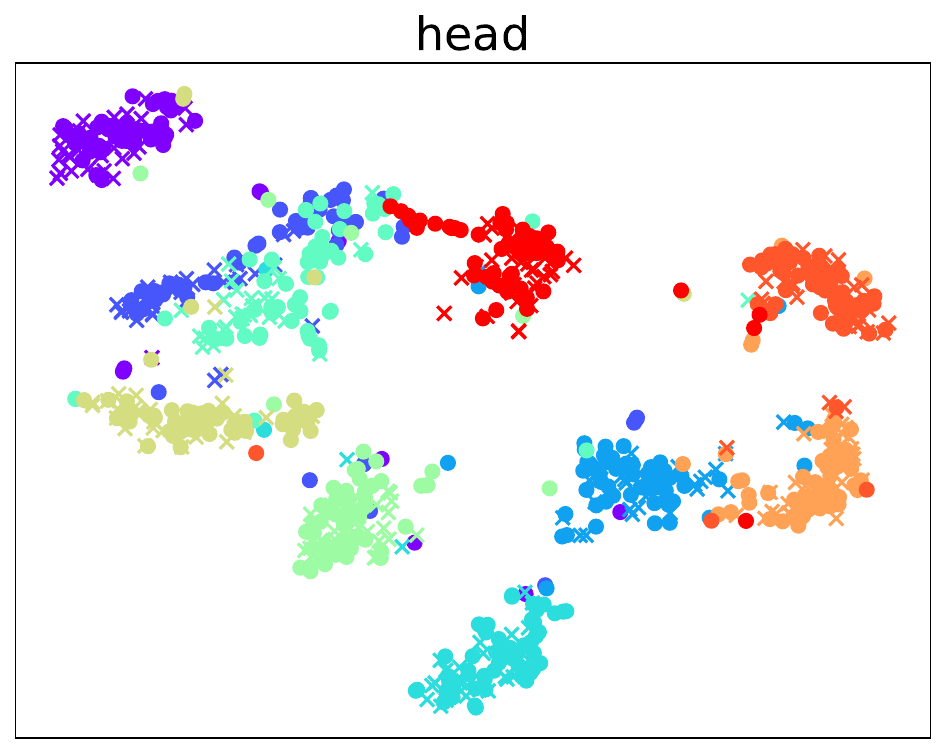}
    \end{subfigure}
    \caption{The t-SNE plots for object features in the artificial triplets. We select 10 classes from each group. Real features are plotted in dots and generated in crosses.}
    \label{fig:tsne}
\end{figure}
\vspace{-2.5mm}

We visualize the similarity between the synthesized object features in artificial triplets and the real ones in the given classes. Fig.~\ref{fig:tsne} illustrates the sampled classes within the body, tail, and head groups, separated by different colors. The neighborhood identity between the real and synthetic features from the same class suggest the effectiveness of the generator.

\section{Conclusion}
In this paper, we introduce two key concepts to enhance the dataset modification approach for unbiased SGG: a novel data augmentation strategy via our FSTA module, and improved predicate reassignment efficiency through Soft Transfer. The FSTA module substantially boosts tail class recall by generating additional artificial triplets, while Soft Transfer offers a more nuanced evaluation of the reliability of individual transfers, allowing for a continuous degree of transfer and mitigating the typical decline in frequent class recall during reassignment. Experimental results confirm that integrating the complementary modules improves overall performance, surpassing the baseline IETrans.

\section*{Acknowledgments}
This work was supported by the commissioned research (No. 225) by National Institute of Information and Communications Technology (NICT), JSPS/MEXT KAKENHI Grant Numbers JP23H03449 and JP22H05015.

\bibliographystyle{ieicetr}
\bibliography{egbib}

\appendix
\setcounter{table}{0}
\renewcommand{\thefigure}{A\arabic{figure}}
\renewcommand{\thetable}{A\arabic{table}}
\renewcommand{\theequation}{A\arabic{equation}}
\renewcommand{\thealgocf}{A\arabic{algocf}}

\section{Full Sgcls and Sgdet Tasks Results}
The full results for the ``sgcls'' and ``sgdet'' tasks are listed in Table \ref{table:sgcls_full} (sgcls) and \ref{table:sgdet_full} (sgdet). 

\section{Qualitative Results}
\begin{figure*}[htb]
    \centering
    \begin{subfigure}[c]{0.3\textwidth}
        \includegraphics[width=\textwidth]{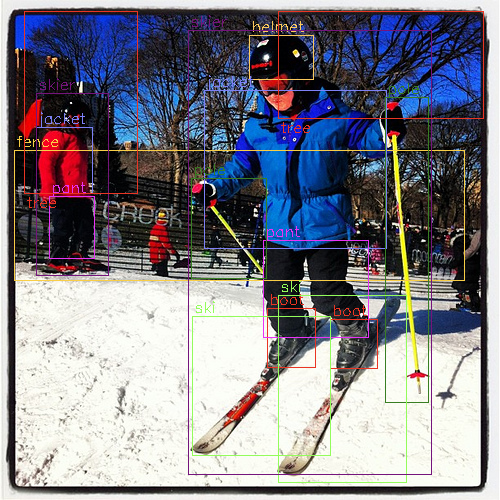}
        \label{fig:sub1}
    \end{subfigure}
    \hfill 
    \begin{subfigure}[c]{0.3\textwidth}
        \includegraphics[trim={6.5cm 3cm 6.5cm 1.8cm},clip,width=\textwidth]{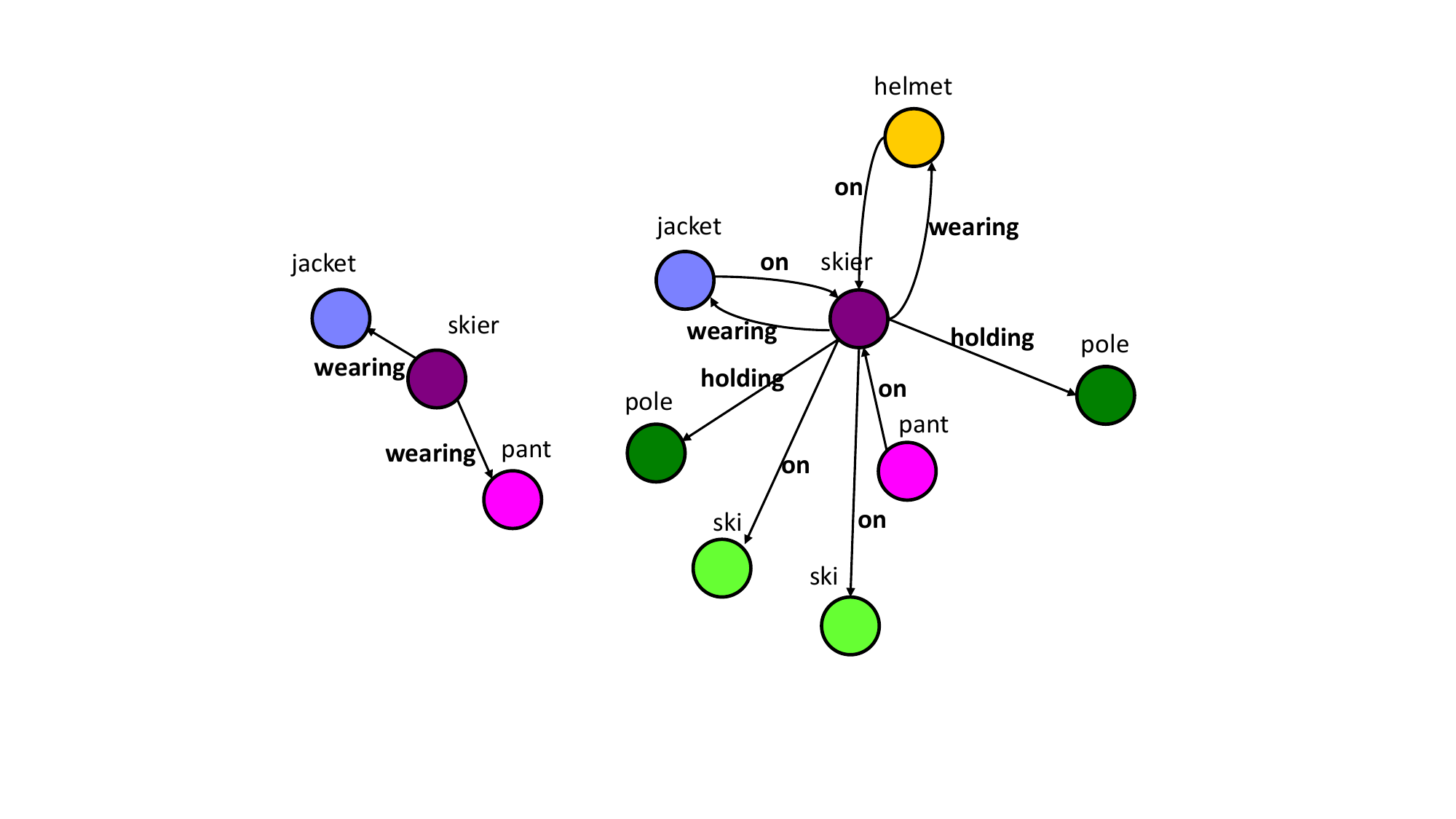}
        \label{fig:sub2}
    \end{subfigure}
    \hfill
    \begin{subfigure}[c]{0.3\textwidth}
        \includegraphics[trim={6.5cm 3cm 6.5cm 1.8cm},clip,width=\textwidth]{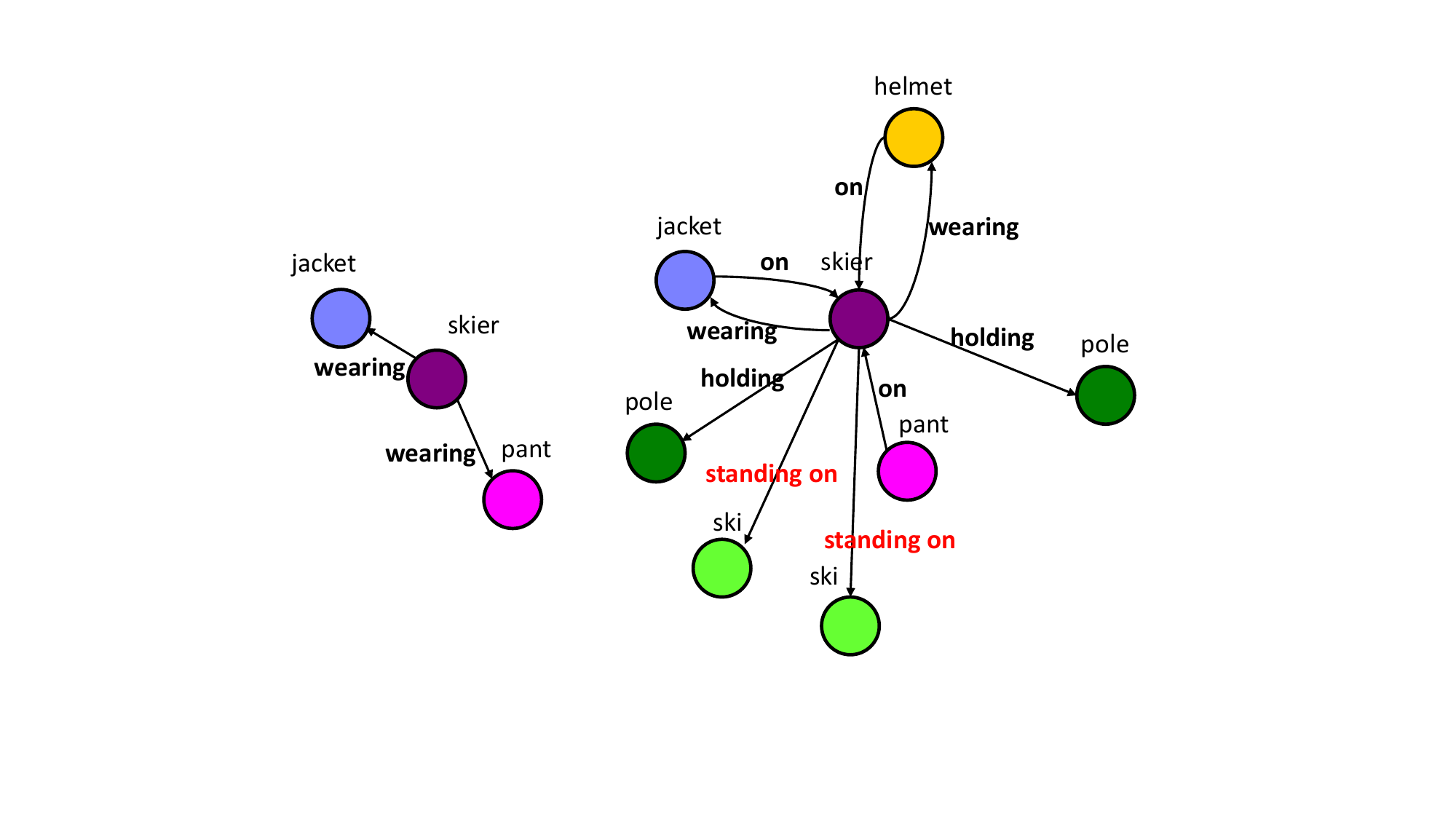}
        \label{fig:sub3}
    \end{subfigure}
    
    \begin{subfigure}[c]{0.3\textwidth}
        \includegraphics[width=\textwidth]{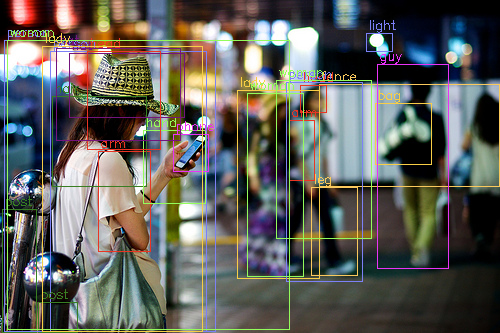}
        \label{fig:sub4}
    \end{subfigure}
    \hfill
    \begin{subfigure}[c]{0.3\textwidth}
        \includegraphics[trim={4.2cm 2.6cm 4.2cm 2.2cm},clip,width=\textwidth]{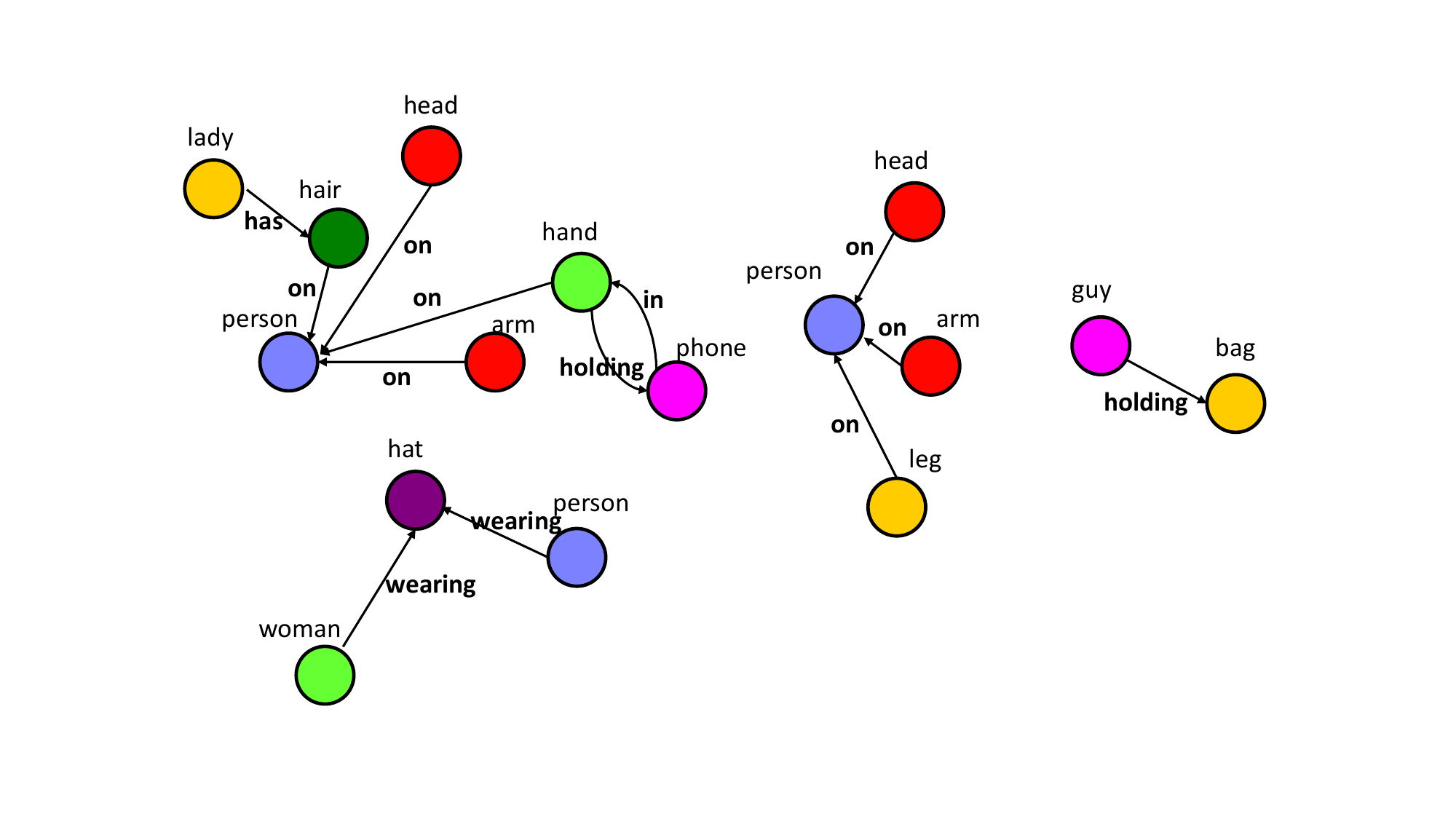}
        \label{fig:sub5}
    \end{subfigure}
    \hfill
    \begin{subfigure}[c]{0.3\textwidth}
        \includegraphics[trim={4.2cm 2.6cm 4.2cm 2.2cm},clip,width=\textwidth]{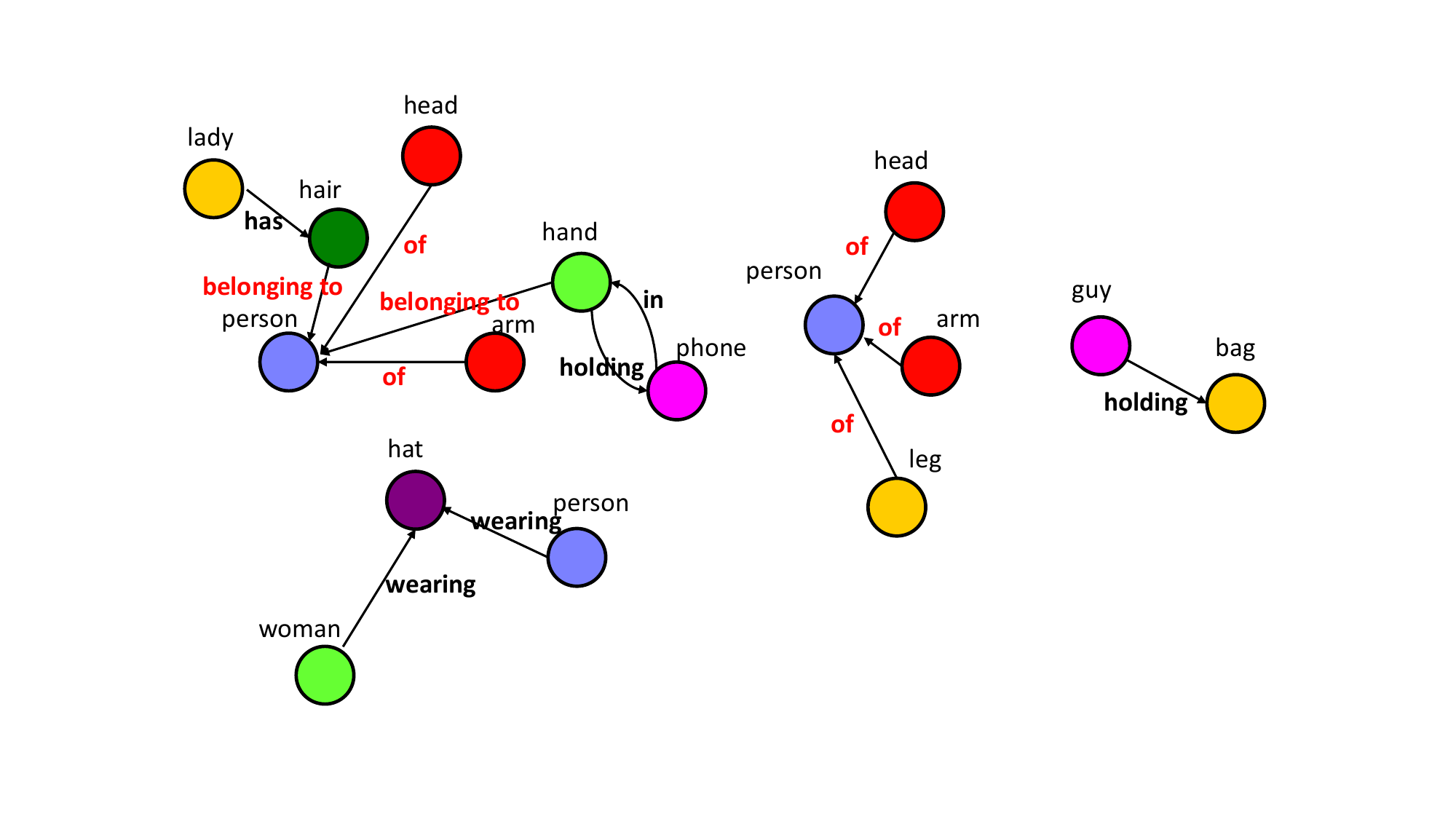}
        \label{fig:sub6}
    \end{subfigure}
    
    \caption{Qualitative results of our method for the predcls task under the Motif+rwt setting: (Left) Images with bounding boxes, (Middle) Ground-truth scene graphs, and (Right) Predicted results. Isolated nodes have been omitted from the visualized scene graphs. The relations in red indicate discrepancies with the ground truth.}
    \label{fig:qual}
\end{figure*}

We visualize the results of predicate prediction in Fig.~\ref{fig:qual}.

\section{Results for Ablation Study (RelDN)}
Table~\ref{tab:abl-reldn} and Table~\ref{tab:abl-reldn-rwt} summarize the ablation results for FSTA module under RelDN and RelDN with reweighting, respectively.

\begin{table}[h]
  \caption{The ablation study results for our FSTA module with RelDN. Item descriptions are identical to Table 4. (Unit: \%)}
  \label{tab:abl-reldn}
  \footnotesize
  \begin{tabular}{cccccc}
    \toprule
    
    us   & +sbj       & MP         & R@100 & mR@100(h/b/t) & F1/Avg@100\\
    \midrule
               &            &            & 38.9 & 33.4(37.0/38.8/24.5) & 35.9/36.2 \\
    \checkmark &            &            & 39.1 & 33.9(37.0/38.2/26.7) & \textbf{36.3}/\textbf{36.5} \\
    \checkmark & \checkmark &            & 39.0 & 34.0(37.0/38.9/26.1) & 36.3/36.5 \\
    \checkmark & \checkmark & \checkmark & 38.2 & 34.1(35.8/36.0/30.5) & 36.0/36.2 \\
    \bottomrule
  \end{tabular}
\end{table}
\vspace{-10mm}
\begin{table}[h]
  \caption{The ablation study results for our FSTA module with RelDN and ``reweighting''. Item descriptions are identical to Table 4. (Unit: \%)}
  \label{tab:abl-reldn-rwt}
  \footnotesize
  \begin{tabular}{cccccc}
    \toprule
    
    us   & +sbj       & MP         & R@100 & mR@100(h/b/t) & F1/Avg@100\\
    \midrule
               &            &            & 25.3 & 34.9(29.9/39.6/35.0) & 29.3/30.1 \\
    \checkmark &            &            & 25.4 & 35.1(30.3/39.4/35.4) & 29.5/30.3 \\
    \checkmark & \checkmark &            & 25.5 & 35.1(30.3/39.8/35.0) & 29.5/30.3 \\
    \checkmark & \checkmark & \checkmark & 25.3 & 35.8(30.1/38.1/38.8) & \textbf{29.6}/\textbf{30.5} \\
    \bottomrule
  \end{tabular}
\end{table}

\section{Results for Sensitivity Analysis (RelDN)}
Table \ref{tab:reldn soft transfer sensitivity} lists the Soft Transfer sensitivity results of RelDN for the predcls task. $k_{s}=0.7$ actually achieves a better score on \textbf{both} R@100 and mR@100 over the baseline method (an improvement of 39.9 to 40.8 for R@100 and 32.3 to 32.5 for mR@100). Nevertheless, modifying $Q(\cdot)=1-Q'(\cdot)$ to $Q(\cdot)=Q'(\cdot)$ leads to stronger overall performance, due to the larger space for the R@100 score recovery. Again, the ``Na\"ive'' case is inferior to the applied settings. 

\begin{table}[ht]
  \caption{The sensitivity results with RelDN for our Soft Transfer module. ``$\ast$'' indicates the setting applied in our method. ``$\diamond$'' stands for a modified $Q(\cdot)$. (Unit: \%)}
  \footnotesize
  \label{tab:reldn soft transfer sensitivity}
  \begin{tabular}{lccc}
    \toprule
    
    settings:         & R@100 & mR@100(h/b/t)        & F1/Avg@100 \\
    \midrule
    $k_{s}=0.5\diamond$ & 50.8  & 29.9(39.5/32.4/18.3) & \textbf{37.6}/40.4 \\
    $k_{s}=0.7\diamond\ast$ & 53.8  & 28.1(40.0/29.7/15.3) & 36.9/\textbf{40.9} \\
    $k_{s}=0.7$     & 40.8  & 32.5(38.0/36.2/23.8) & 36.2/36.7 \\
    Na\"ive         & 56.3  & 25.0(40.9/25.5/9.6)  & 34.7/40.7 \\ 
    \bottomrule
  \end{tabular}
\end{table}

\section{Results for FSTA Parameter Choices (RelDN)}

Table~\ref{tab:sensi-reldn-rwt} describes the results of FSTA parameter study for RelDN.

\begin{table}[h]
  \caption{The results of parameter choices for FSTA with RelDN and ``reweighting''. ``$\ast$'' indicates the setting applied in our method. (Unit: \%)}
  \footnotesize
  \label{tab:sensi-reldn-rwt}
  \begin{tabular}{ll|ccc}
    \toprule
    
    Param. & Value    & R@100 & mR@100(h/b/t)        & F1/Avg@100 \\
    \midrule
    \multirow{5}{*}{$s_{iou}$} & 0.5$\ast$   & 25.3 & 35.8(30.1/38.1/38.8) & \textbf{29.6}/\textbf{30.5} \\ 
    &0.6       & 24.8 & 35.0(29.5/38.2/36.9) & 29.0/29.9 \\
    &0.7       & 25.2 & 34.4(29.4/38.5/35.0) & 29.1/29.8 \\ 
    &0.8       & 24.3 & 34.0(29.2/36.8/35.7) & 28.3/29.2 \\
    &0.9       & 24.5 & 36.3(29.3/37.8/41.4) & \underline{29.3}/\underline{30.4} \\
    \midrule
    \multirow{5}{*}{$U_h$} & 0.2       & 24.6 & 35.7(29.5/38.1/39.0) & 29.1/30.2 \\ 
    &0.4       & 24.3 & 36.8(29.1/37.5/43.3) & 29.3/\underline{30.5} \\
    &0.5       & 25.4 & 36.1(30.0/37.7/40.1) & \textbf{29.8}/\textbf{30.8} \\ 
    &0.6       & 24.5 & 35.8(29.3/38.1/39.5) & 29.1/30.2 \\
    &0.8$\ast$   & 25.3 & 35.8(30.1/38.1/38.8) & \underline{29.6}/\underline{30.5} \\
    \bottomrule
  \end{tabular}
\end{table}

\section{Randomness of FSTA}
\noindent{\textbf{The resource of randomness}}: Including the undersampling step and the generator pretraining. We selected a fixed checkpoint for the generator based on the classification accuracies observed in the validation data.

\noindent{\textbf{The reproducibility of randomness}}: In the SGG model training, we followed the open-source SGG model implementation\ref{fn1} to set the seed for the libraries and switch to deterministic mode for the cudnn library.

\noindent{\textbf{The impact of randomness}}: We measured the standard deviation of R@100 and mR@100 under ``Motif++FSTA+rwt'' in the predcls task, using five different runs. The values are 0.23 for R@100 and 0.31 for mR@100. Note that these include randomness from both the Motif model and the FSTA module.

\begin{table*}[ht]
  \caption{The full performance comparison for the sgcls task on VG150. Scores for models listed in the first section are cited from their original papers, while models in subsequent sections use our implementation. ``Model++X'' is shorthand for ``Model+IETrans+X''. The best overall scores within each section are highlighted in bold. (Unit: \%) 
  }
  \footnotesize
  \label{table:sgcls_full}
  \begin{tabular}{lcccccccc}
    \toprule
           &  \multicolumn{8}{c}{Scene Graph Classification (Sgcls)} \\
    \cmidrule(lr){2-9}
    models &R@50 & R@100 & mR@50(h/b/t) & mR@100(h/b/t) & F1@50 & F1@100 & A@50 & A@100\\
    \midrule
    Motif+TDE \cite{TDE} & 27.7 & 29.9 & 13.1(-) & 14.9(-) & 17.8 & 19.9 & 20.4 & 22.4\\
    Motif+DLFE \cite{DLFE} & 32.3 & 33.1 & 15.2(-) & 15.9(-) & 20.7 & 21.5 & 23.8& 24.5\\
    Motif+NICE \cite{NICE}  & 33.1 & 34.0 & 16.6(-) & 17.9(-) & 22.1 & 23.5 & 24.9& 26.0\\
    Motif+IETrans \cite{IETrans} & 32.5 & 33.4 & 16.8(-) & 17.9(-) & 22.2 & 23.3 & 24.7 & 25.7\\
    Motif+IETrans+rwt \cite{IETrans} & 29.4 & 30.2 & 21.5(-) & 22.8(-) & 24.8 & 26.0 & 25.5 & 26.5\\
    Motif+Inf \cite{inf_CVPR23} & 32.2 & 33.8 & 14.5(-) & 17.4(-) & 20.0 & 23.0 & 23.4 & 25.6 \\
    \midrule
    Motif                  & 38.1 & 38.9 & 10.0(23.9/6.3/0.6) & 10.7(25.2/7.0/0.8) &  15.8 & 16.8 & 24.1 & 24.8\\
    Motif+IETrans\dag      & 29.1 & 30.1 & 18.0(24.5/19.7/10.3) & 20.9(25.9/21.2/15.9) &  22.2 & \textbf{24.7} & 23.6 & 25.5\\
    Motif++FSTA (ours)      & 29.5 & 30.5 & 18.3(24.5/19.4/11.5) & 20.6(26.0/21.0/15.1) & 22.6 & 24.6 & 23.9 & 25.6\\
    Motif++SoftTrans (ours) & 33.0 & 34.1 & 17.2(24.9/19.1/8.1) & 18.7(26.4/20.8/9.3) & 22.6 & 24.2 & \textbf{25.1} & \textbf{26.4}\\
    Motif++Full (ours)       & 32.2 & 33.3 & 17.7(24.2/18.7/10.5) & 19.2(25.7/20.5/11.8) & \textbf{22.8} & 24.4 & 25.0 & 26.3\\
    \midrule
    Motif+IETrans+rwt\dag & 28.1 & 28.6 & 18.9(24.1/20.2/12.6) & 21.0(25.0/20.9/17.4) & 22.6 & 24.2 & 23.5 & 24.8 \\
    Motif++FSTA+rwt (ours) & 26.1 & 26.6 & 19.6(23.3/20.2/15.4) & 21.6(24.2/20.9/20.0) & 22.4 & 23.8 & 22.9 & 24.1 \\
    Motif++SoftTrans+rwt (ours) & 30.9 & 31.5 & 18.0(24.2/19.2/10.9) & 20.4(25.0/20.0/16.3) & 22.7 & \textbf{24.8} & \textbf{24.5} & \textbf{26.0}\\
    Motif++Full+rwt (ours) & 29.3 & 29.9 & 18.5(23.6/19.3/13.1) & 21.0(24.4/20.0/18.7) & \textbf{22.7} & 24.7 & 23.9 & 25.5 \\
    \midrule
    \midrule
    RelDN                  & 36.0 & 36.9 & 7.4(21.8/1.3/0.0) & 7.9(22.9/1.6/0.0) & 12.3 & 13.0 & 21.7 & 22.4\\
    RelDN+IETrans\dag      & 22.4 & 23.3 & 17.8(20.8/20.0/12.7) & 19.0(22.0/21.2/14.1) & 19.8 & 20.9 & 20.1 & 21.2\\
    RelDN++FSTA (ours)      & 21.9 & 22.8 & 17.9(20.4/19.9/13.6) & 19.3(21.6/21.2/15.2) & 19.7 & 20.9 & 19.9 & 21.1\\
    RelDN++SoftTrans (ours) & 31.8 & 32.8 & 14.6(22.2/15.9/6.2) & 15.5(23.4/16.9/6.7) & 20.0 & 21.1 & \textbf{23.2} & \textbf{24.2}\\ 
    RelDN++Full (ours)  & 30.1 & 31.1 & 15.6(21.7/16.3/9.4) & 16.8(22.9/17.4/10.6) & \textbf{20.5} & \textbf{21.8} & 22.9 & 24.0\\ 
    \midrule
    RelDN+IETrans+rwt\dag & 17.7 & 18.5 & 19.3(18.7/21.0/18.1) & 20.6(19.9/22.1/19.7) & 18.5& 19.5& 18.5& 19.6\\
    RelDN++FSTA+rwt (ours) & 16.6& 17.4& 19.0(18.0/21.4/17.4)& 20.8(19.1/22.5/20.5)& 17.7& 18.9&17.8 & 19.1\\
    RelDN++SoftTrans+rwt (ours) & 23.7& 24.6& 18.4(21.6/18.6/15.3)& 21.1(22.8/19.5/21.0) & \textbf{20.7} & \textbf{22.7}&\textbf{21.0} & \textbf{22.9}\\
    RelDN++Full+rwt (ours) & 22.6& 23.5& 18.7(20.6/19.0/16.5)& 21.6(21.8/20.0/23.1)& 20.5& 22.5& 20.7&22.6 \\
    \bottomrule
  \end{tabular}
\end{table*}

\begin{table*}[ht]
  \caption{The full performance comparison for the sgdet task on VG150. Scores for models listed in the first section are cited from their original papers, while models in subsequent sections use our implementation. ``Model++X'' is shorthand for ``Model+IETrans+X''. The best overall scores within each section are highlighted in bold. (Unit: \%) 
  }
  \footnotesize
  \label{table:sgdet_full}
  \begin{tabular}{lcccccccc}
    \toprule
           &  \multicolumn{8}{c}{Scene Graph Detection (Sgdet)} \\
    \cmidrule(lr){2-9}
    models &R@50 & R@100 & mR@50(h/b/t) & mR@100(h/b/t) & F1@50 & F1@100 & A@50 & A@100\\
    \midrule
    Motif+TDE \cite{TDE} & 16.9 & 20.3 & 8.2(-) & 9.8(-) & 11.0 & 13.2 & 12.5 & 15.1\\
    Motif+DLFE \cite{DLFE} & 25.4 & 29.4 & 11.7(-) & 13.8(-) & 16.0 & 18.8 & 18.6 & 21.6\\
    Motif+NICE \cite{NICE}  & 27.8 & 31.8 & 12.2(-) & 14.4(-) & 17.0 & 19.8 & 20.0 & 23.1\\
    Motif+IETrans \cite{IETrans} & 26.4 & 30.6 & 12.4(-) & 14.9(-) & 16.9 & 20.0 & 19.4 & 22.8\\
    Motif+IETrans+rwt \cite{IETrans} & 23.5 & 27.2 & 15.5(-) & 18.0(-) & 18.7 & 21.7 & 19.5 & 22.6\\
    Motif+Inf \cite{inf_CVPR23} & 23.9 & 27.1 & 9.4(-) & 11.7(-) & 13.5  & 16.3 & 16.7 & 19.4 \\
    \midrule
    Motif                  & 32.7 & 37.7 & 7.7(19.4/4.4/0.1) & 9.3(23.0/5.6/0.2) & 12.5 & 14.9 & 20.2 & 23.5\\
    Motif+IETrans\dag      & 24.7 & 29.2 & 13.8(20.9/15.5/5.3) & 16.5(24.9/18.4/6.8) & 17.7 & 21.1 & 19.3 & 22.9\\
    Motif++FSTA (ours)      & 24.3 & 28.8 & 13.9(21.2/15.7/5.3) & 17.1(25.1/18.3/8.2) & 17.7 & 21.5 & 19.1 & 23.0\\
    Motif++SoftTrans (ours) & 27.4 & 32.2 & 13.1(21.1/15.5/3.2) & 15.8(25.2/18.6/4.1) & 17.7 & 21.2 & 20.3 & 24.0\\
    Motif++Full (ours)      & 27.5 & 32.2 & 14.0(20.8/15.7/5.8) & 17.0(24.6/18.5/8.3) & \textbf{18.6} & \textbf{22.3}& \textbf{20.8} & \textbf{24.6}\\
    \midrule
    Motif+IETrans+rwt\dag & 23.8 & 28.5 & 15.6(22.5/18.3/6.3) & 18.8(26.5/21.1/9.4) & 18.8 & 22.7 & 19.7 & 23.7 \\
    Motif++FSTA+rwt (ours) & 22.1 & 26.4 & 17.3(21.3/18.7/12.0) & 20.1(25.2/21.1/14.2) & 19.4 & 22.8 & 19.7 & 23.3 \\
    Motif++SoftTrans+rwt (ours) & 27.0 & 31.8 & 15.0(22.1/17.6/5.8) & 19.4(26.1/20.4/12.0) & 19.3 & \textbf{24.1} & \textbf{21.0} & \textbf{25.6}\\
    Motif++Full+rwt (ours) & 25.5 & 30.1 & 16.3(21.4/18.2/9.5) & 19.5(25.2/20.8/12.9) & \textbf{19.9} & 23.7 & 20.9 & 24.8 \\
   
    \midrule
    \midrule
    RelDN                  & 32.7 & 38.0 & 6.7(19.7/1.1/0.0) & 8.2(23.7/1.9/0.0) & 11.1 & 13.5 & 19.7 & 23.1\\
    RelDN+IETrans\dag      & 18.4 & 22.0 & 14.9(18.3/16.8/9.7) & 18.4(22.0/20.9/12.4) & 16.5 & 20.0 & 16.7 & 20.2\\
    RelDN++FSTA (ours)      & 17.5 & 21.1 & 15.5(17.8/17.0/11.7) & 19.1(21.3/21.0/15.1) & 16.4 & 20.1 & 16.5 & 20.1\\
    RelDN++SoftTrans (ours) & 27.3 & 32.3 & 12.5(19.9/13.3/4.7) & 15.4(23.8/16.5/6.3) & 17.1 & 20.9 & \textbf{19.9} & \textbf{23.9}\\ 
    RelDN++Full (ours)  & 24.6 & 29.3 & 14.4(19.0/14.7/9.8) & 17.2(22.5/17.8/11.5) & \textbf{18.2} & \textbf{21.7} & 19.5 & 23.3\\
    \midrule
    RelDN+IETrans+rwt\dag & 12.2 & 14.7 & 16.5(14.9/19.8/14.8) & 19.7(17.8/23.2/17.9) & 14.0 & 16.8 & 14.4 & 17.2 \\
    RelDN++FSTA+rwt (ours) & 11.2 & 13.8 & 16.6(14.8/19.0/15.8) & 19.5(17.7/22.5/18.2) & 13.4 & 16.2 & 13.9 & 16.7\\
    RelDN++SoftTrans+rwt (ours) & 18.0 & 21.6 & 15.9(18.6/18.3/11.1) & 18.9(22.1/21.5/13.5) & \textbf{16.9} & \textbf{20.2} & \textbf{17.0} & \textbf{20.3} \\
    RelDN++Full+rwt (ours) & 16.0 & 19.5 & 16.5(17.3/18.0/14.2) & 19.9(20.5/21.4/17.7) & 16.2 & 19.7 & 16.3 & 19.7\\
    \bottomrule
  \end{tabular}
\end{table*}

\section{Object Generator}

We exploit a conditional-GAN based model to synthesize  \verb+object'+ features, due to its lightweight and low additional computational cost. In the pre-processing step, we collect the real features from model predictions on training data (See Fig.4 in the manuscript). The adversarial loss function for the GAN model consist of three parts: $\mathcal{L}_{wgangp}$, $\mathcal{L}_{cls}$, and $\mathcal{L}_{recon}$. 

$\mathcal{L}_{wgangp}$ is a standard WGAN loss with gradient penalty \cite{wgangp} as Eq.(\ref{eq:loss_wgangp}). 

\begin{equation}
  \begin{aligned}
    \mathcal{L}_{wgangp} = &\mathbb{E}_{\mathbf{x} \sim real}[D(\mathbf{x}, \mathbf{s}_{c})] - \mathbb{E}_{\tilde{\mathbf{x}} \sim gen}[D(\tilde{\mathbf{x}}, \mathbf{s}_{c})] \\ &-\lambda \mathbb{E}[(||\nabla_{\hat{\mathbf{x}}}D(\hat{\mathbf{x}}, \mathbf{s}_{c})||_{2} -1)^{2}]
  \end{aligned}
  \label{eq:loss_wgangp}
\end{equation}
where $\mathbf{x} \in \mathbb{R}^{d}$ is the feature sampled from real data, $\tilde{\mathbf{x}} = G(\mathbf{z}, \mathbf{s}_{c}) \in \mathbb{R}^{d}$ is the synthesized feature from generator $G$. $d$ is the size of object feature. $\hat{\mathbf{x}} = \alpha \mathbf{x} + (1 - \alpha) \tilde{\mathbf{x}}$ is an interpolated feature with $\alpha$ sampled from a uniform distribution. $\mathbf{z}$ is an initial vector sampled from normal distribution, and $\mathbf{s}_{c}$ is a condition vector represents the object class. We collect $\mathbf{s}_{c}$ from the pre-trained CLIP \cite{CLIP} text encoder. We use the basic template ``a photo of a [OBJECT NAME].'' as the input prompt to text encoder, then the output vector as the class representation.

$\mathcal{L}_{cls}$ is a regularization loss for the generator $G$. It utilizes a softmax classifier pre-trained on real data to encourage the generator to output features with enhanced discriminability. That is, the synthetic features can be better classified. Eq.(\ref{eq:cls}) describes its loss function.

\begin{equation}
    \mathcal{L}_{cls} = - \mathbb{E}_{\tilde{\mathbf{x}} \sim gen}[\mathrm{log}  P(y|\tilde{x}; \theta_{cls})]
    \label{eq:cls}
\end{equation}
where $\theta_{cls}$ is the weights of the softmax classifier. $y$ is the corresponding class label. During the adversarial training, the pre-trained classifier is frozen.

$\mathcal{L}_{recon}$ is another regularization term for the class consistency between generator output and its condition input. A reconstructor $R(\cdot)$ is pre-trained on real data to infer the class condition vector from the feature. Eq.(\ref{eq:recon}) describes its loss function.

\begin{equation}   
    \mathcal{L}_{recon} = \mathbb{E}_{\tilde{\mathbf{x}} \sim gen}[\lVert R(\tilde{\mathbf{x}}) - \mathbf{s}_{c} \rVert_{2}]
    \label{eq:recon}
\end{equation}
the reconstructor is also frozen in the adversarial training.

The overall loss function is as below and identical to Eq.(4) in the main paper.

\begin{equation} 
    \min_{G}\max_{D} \mathcal{L}_{wgangp} + \beta \mathcal{L}_{cls} + \gamma \mathcal{L}_{recon}
    \label{eq:loss_all}
\end{equation}

We list the model architecture for training the object generator in Table \ref{table:gan-arch}.

\begin{table}[ht]
  \caption{The model architecture.}
  \footnotesize
  \label{table:gan-arch}
  \centering
  \begin{tabular}{ll}
    \toprule
    
    module     & input and the forward flow      \\
    \midrule
    $G$ & \textbf{Input: ($\mathbf{z}$, $\mathbf{s_{c}}$)} \\
             & out = concat(Input) \\
             & out = linear1(in=1024+512, out=4096)(out) \\ 
             & out = LeakyReLU(slope=-0.2)(out)  \\
             & $\tilde{\mathbf{x}}$ = linear2(in=4096, out=1024)(out)  \\
    \midrule
    $D$ & \textbf{Input: ($\tilde{\mathbf{x}}$, $\mathbf{s_{c}}$) or ($\mathbf{x}$, $\mathbf{s_{c}}$) } \\
             & out = concat(Input) \\
             & out = linear1(in=1024+512, out=4096)(out) \\ 
             & out = LeakyReLU(slope=-0.2)(out)  \\
             & out = linear2(in=4096, out=1)(out)  \\
    \midrule
    classifier & \textbf{Input: $\tilde{\mathbf{x}}$} \\
                   & out = linear(in=1024, out=150)(out) \\
                   & out = softmax(out) \\
    \midrule
    reconstructor & \textbf{Input: $\tilde{\mathbf{x}}$} \\
                   & out = linear(in=1024, out=4096)(out) \\
                   & out = LeakyReLU(slope=-0.2)(out)  \\
                   & out = linear(in=4096, out=512)(out) \\             
    \bottomrule
  \end{tabular}
\end{table}

\section{Hyperparameter Details}

We list the parameter choices for training SGG models and the generator model in Table \ref{table:SGG}.

\begin{table}[ht]
  \caption{The parameter choices for training Motif-based SGG models (section 1), RelDN-based SGG models (section 2), and the genertor model (section 3).}
  \footnotesize
  \label{table:SGG}
  \centering
  \begin{tabular}{lll}
    \toprule
    
    Parameter     & Value  & Description      \\
    \midrule
    MOTIF\_IMS\_PER\_BATCH  & 16 & batch size \\
    MOTIF\_BASE\_LR & 0.015 & learning rate \\
    MOTIF\_MAX\_ITER & 40,000 & iterations \\
    \midrule
    RELDN\_IMS\_PER\_BATCH  & 2 & batch size \\
    RELDN\_BASE\_LR & 0.005 & learning rate \\
    RELDN\_MAX\_ITER & 150,000 & iterations \\
    \midrule    
    $d_{z}$ & 1024 & dim of input $z$\\
    BATCH\_FG  &  128  &  batch size (adv. training) \\
    D\_TRAIN\_ITER & 5 & D-over-G update iters \\
    MAX\_ITER\_FG & 55,000 & iterations \\
    GAN\_LR  & 0.0001 & learning rate\\
    $\lambda$ & 10.0 & the coef. for gp \\ 
    $\beta$ & 0.1 & the coef. for loss $\mathcal{L}_{cls}$\\
    $\gamma$ & 0.1 & the coef. for loss $\mathcal{L}_{recon}$\\
    
    \bottomrule
  \end{tabular}
\end{table}

\end{document}